%% file: star.tex
\documentclass[10pt,journal,compsoc]{IEEEtran}
\ifCLASSOPTIONcompsoc
  \usepackage[nocompress]{cite}
\else
  \usepackage{cite}
\fi
\ifCLASSINFOpdf
\else
\fi

\usepackage{graphicx}                

\graphicspath{{images/}}

\usepackage{amsmath, amsthm, amsfonts, amssymb}
\usepackage{mathrsfs}
\usepackage{microtype}                 
\PassOptionsToPackage{warn}{textcomp}  
\usepackage{textcomp}                  
\usepackage{mathptmx}                  
\usepackage{times}                     
\usepackage{cite}                      
\usepackage{tabu}                      
\usepackage{booktabs}                  
\usepackage[normalem]{ulem}
\usepackage{color}
\usepackage{algorithm}
\usepackage{algpseudocode}
\usepackage{microtype}
\usepackage{multicol}
\usepackage{ulem}

\newif\ifshowcomments
\showcommentsfalse

\definecolor{turquoise}{cmyk}{0.65,0,0.1,0.1}
\definecolor{purple}{rgb}{0.65,0,0.65}
\definecolor{dark_green}{rgb}{0, 0.5, 0}
\definecolor{orange}{rgb}{0.8, 0.6, 0.2}
\definecolor{red}{rgb}{0.8, 0.2, 0.2}
\definecolor{brown}{rgb}{0.5, 0.16, 0.16}
\definecolor{gold}{rgb}{0.85, 0.65 ,0.125}

\ifshowcomments

\newcommand{\rh}[1]{{\color{blue}{#1}}}

\newcommand{\od}[1]{{\color{blue}{#1}}}

\newcommand{\change}[1]{{\color{black} #1}}

\newcommand{\ok}[1]{{\color{red}{[OK: #1]}}}

\else

\newcommand{\rh}[1]{{\color{black} #1}}
\newcommand{\od}[1]{{\color{black} #1}}
\newcommand{\change}[1]{{\color{black} #1}}
\newcommand{\ok}[1]{{\color{black} #1}}

\fi

\renewcommand{\paragraph}[1]{\textit{#1}}

\begin{document}

\title{Shape-driven Coordinate Ordering for Star Glyph Sets via Reinforcement Learning}

\author{
	Ruizhen~Hu, Bin~Chen, Juzhan~Xu, Oliver~van~Kaick, Oliver~Deussen, and~Hui~Huang%
	\IEEEcompsocitemizethanks{
		\IEEEcompsocthanksitem Ruizhen~Hu, Bin~Chen, Juzhan~Xu, and~Hui~Huang are with College of Computer Science and Software Engineering, Shenzhen University. E-mail: \{ruizhen.hu, codeb.box, juzhan.xu, hhzhiyan\}@gmail.com
		\IEEEcompsocthanksitem Oliver~van~Kaick is with the School of Computer Science, Carleton University. E-mail: Oliver.vanKaick@carleton.ca
		\IEEEcompsocthanksitem Oliver~Deussen is with Department of Computer and Information Science, University of Konstanz, and SIAT Shenzhen. E-mail: oliver.deussen@uni-konstanz.de
		\IEEEcompsocthanksitem Hui~Huang is the corresponding author of this paper.
	}%
}

%



\IEEEtitleabstractindextext{%
\begin{abstract}
\input{abstract}
\end{abstract}

}

\maketitle

\IEEEdisplaynontitleabstractindextext

%
\IEEEpeerreviewmaketitle

\begin{figure*}[!t]
	\centering
	\includegraphics[width=0.95\linewidth]{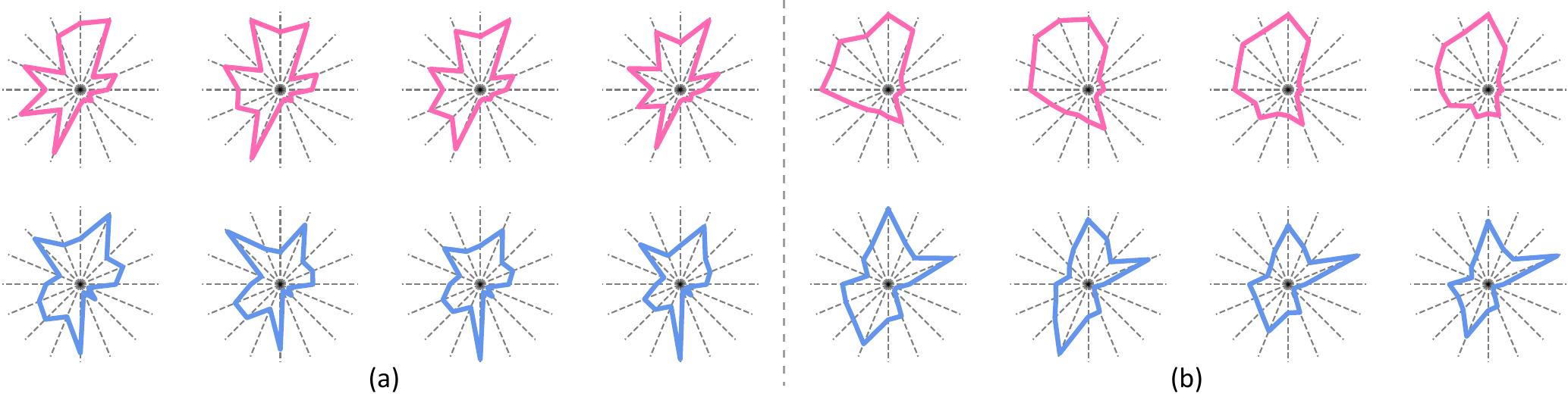}
	\caption{Ordering axes for a set of star glyphs using reinforcement learning. For the same set of high-dimensional data points, we can optimize the axis order of the corresponding star glyphs according to different perceptual criteria: (a) spike and salient shape strategies that place dissimilar axes close-by; (b) maximizing class separability. The class labels of the glyphs are indicated by blue/red color. Star glyphs of the same data point (but with different axis orders) are drawn at the same position inside each plot for comparison.}
	\label{fig:teaser}
\end{figure*}

\input{introduction}
\input{related}

\input{method}

\input{results}
\input{conclusion}

\section*{Acknowledgment}
We thank the anonymous reviewers for their valuable comments. This work was supported in part by NSFC (61872250, U2001206, 61902254), GD Talent Program (2019JC05X328), GD Science and Technology Program (2020A0505100064, 2015A030312015), DEGP Key Project (2018KZDXM058), Shenzhen Innovation Program (JCYJ20180305125709986), NSERC (2015-05407), DFG (422037984), National Engineering Laboratory for Big Data System Computing Technology, and Guangdong Laboratory of Artificial Intelligence and Digital Economy (SZ).

%
%


\bibliographystyle{IEEEtran}
\bibliography{IEEEabrv,star}

\begin{IEEEbiography}[{\includegraphics[width=1in,height=1.25in,clip,keepaspectratio]{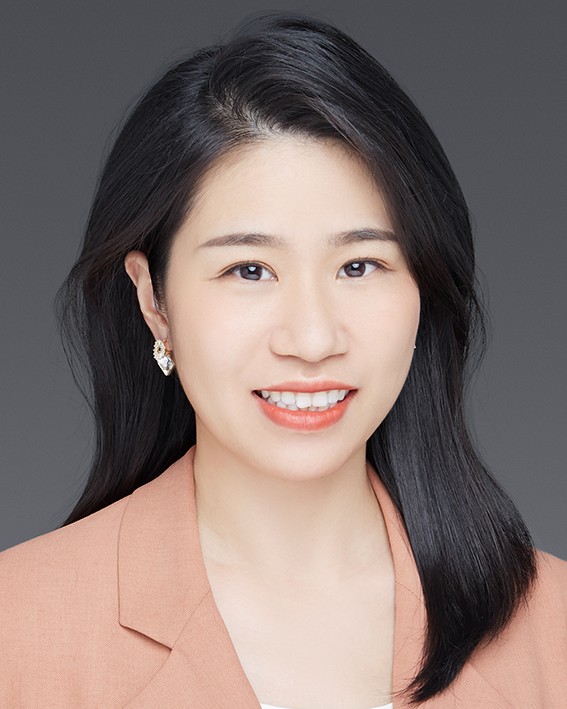}}]{Ruizhen Hu}
is an Associate Professor at Shenzhen University, China. She received her Ph.D. from the Department of Mathematics, Zhejiang University. Her research interests are in computer graphics, with a recent focus on applying machine learning to advance the understanding and generative modeling of visual data including 3D shapes and indoor scenes. She is an editorial board member of The Visual Computer and IEEE CG\&A.
\end{IEEEbiography}
\vspace{-10 mm}

\begin{IEEEbiography}[{\includegraphics[width=1in,height=1.25in,clip,keepaspectratio]{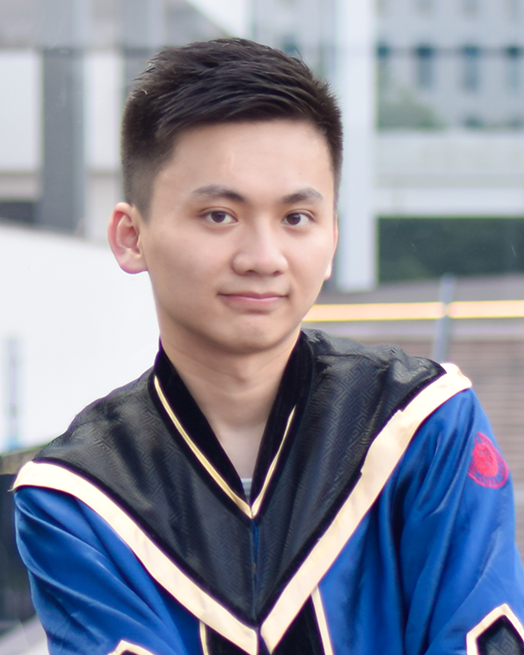}}]{Bin Chen}
	received the bachelor degree in computer science from Shenzhen University in 2019. He is currently working toward the Master degree in Shenzhen University. His research interest include computer graphics and visualization.
\end{IEEEbiography}
\vspace{-10 mm}

\begin{IEEEbiography}[{\includegraphics[width=1in,height=1.25in,clip,keepaspectratio]{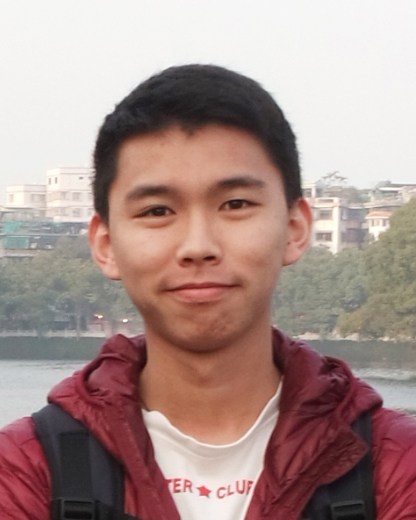}}]{Juzhan Xu}
	received the bachelor degree in computer science from Shenzhen University in 2019. He is currently working toward the Master degree in Shenzhen University. His research interest include computer graphics and visualization.
\end{IEEEbiography}
\vspace{-10 mm}

\begin{IEEEbiography}[{\includegraphics[width=1in,height=1.25in,clip,keepaspectratio]{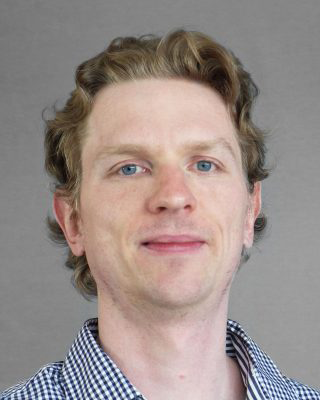}}]{Oliver~van~Kaick}
is an Associate Professor in the School of Computer Science at Carleton University, Ottawa, Canada. He received a Ph.D. from the School of Computing Science at Simon Fraser University (SFU). Oliver was then a postdoctoral researcher at SFU and Tel Aviv University. Oliver's research is concentrated in shape analysis and geometric modeling.
\end{IEEEbiography}
\vspace{-10 mm}

\begin{IEEEbiography}[{\includegraphics[width=1in,height=1.25in,clip,keepaspectratio]{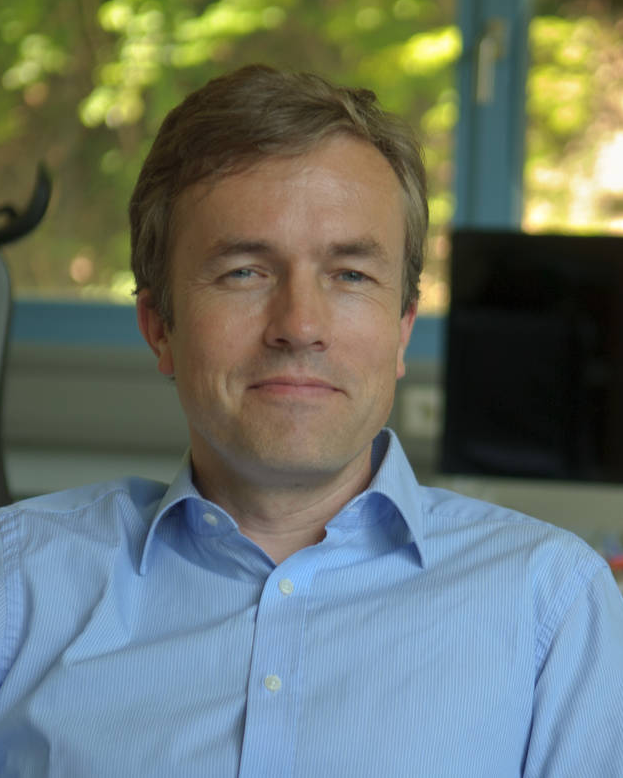}}]{Oliver Deussen}
is a Professor for computer graphics and media informatics with the University of Konstanz. He received his PhD from KIT (Karlsruhe Institute of Technology, Germany). His research interests cover modeling problems, non-photorealistic computer graphics and information visualization. He served as President of the Eurographics Association from 2018-2020. He is a member of ACM Siggraph, Eurographics, and Gesellschaft fuer Informatik.
\end{IEEEbiography}
\vspace{-10 mm}

\begin{IEEEbiography}[{\includegraphics[width=1in,height=1.25in,clip,keepaspectratio]{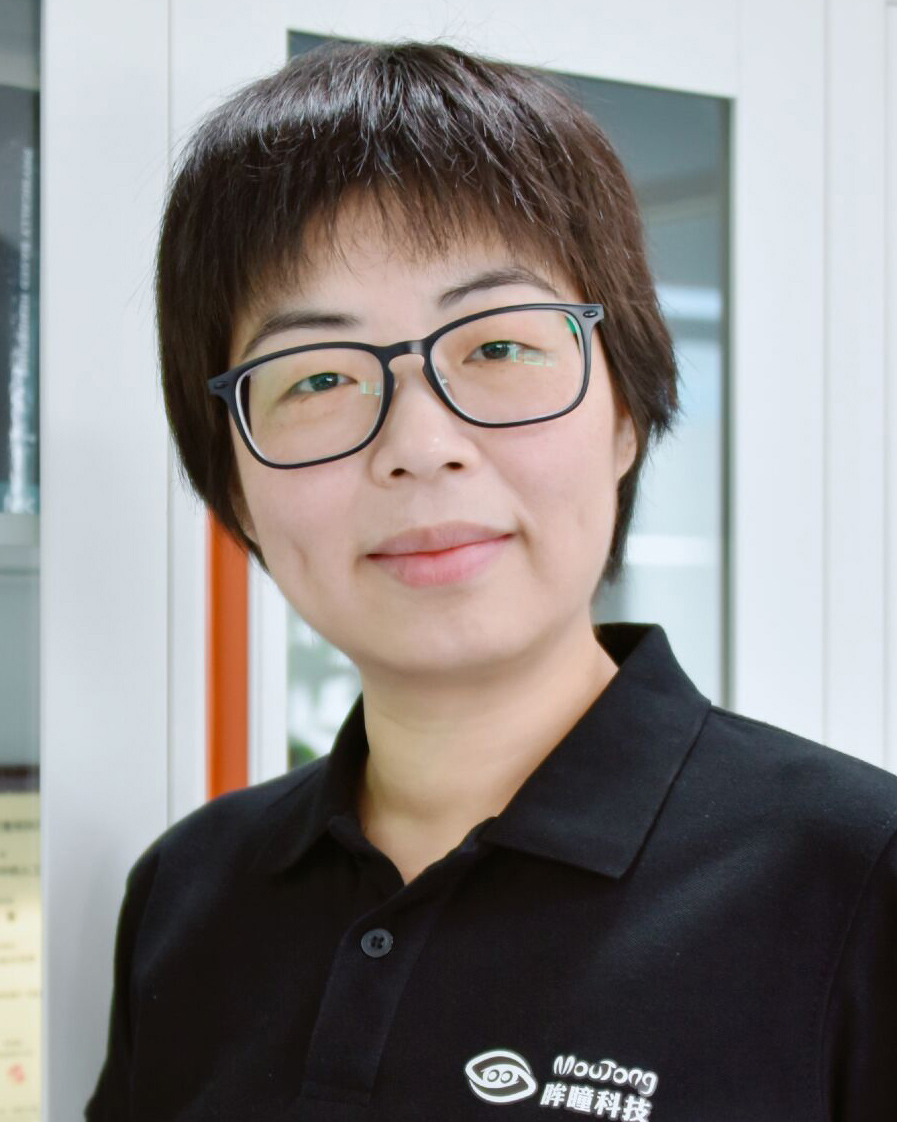}}]{Hui~Huang}
is a Distinguished TFA Professor of Shenzhen University, where she directs the Visual Computing Research Center. She received her PhD in Applied Math from The University of British Columbia in 2008. Her research interests span on Computer Graphics, Vision and Visualization. She is currently a Senior Member of IEEE/ACM/CSIG, a Distinguished Member of CCF, and is on the editorial board of ACM Trans. on Graphics and Computers $\&$ Graphics.
\end{IEEEbiography}







\end{document}

%% file: abstract.tex
We present a neural optimization model trained with reinforcement learning to solve the coordinate ordering problem for sets of star glyphs. Given a set of star glyphs associated to multiple class labels, we propose to use shape context descriptors to measure the perceptual distance between pairs of glyphs, and use the derived silhouette coefficient to measure the perception of class separability within the entire set. To find the optimal coordinate order for the given set, we train a neural network using reinforcement learning to reward orderings with high silhouette coefficients. The network consists of an encoder and a decoder with an attention mechanism. The encoder employs a recurrent neural network (RNN) to encode input shape and class information, while the decoder together with the attention mechanism employs another RNN to output a sequence with the new coordinate order. 
In addition, we introduce a neural network to efficiently estimate the similarity between shape context descriptors, which allows to speed up the computation of silhouette coefficients and thus the training of the axis ordering network.
Two user studies demonstrate that the orders provided by our method are preferred by users for perceiving class separation. 
We tested our model on different settings to show its robustness and generalization abilities and demonstrate that it allows to order input sets with unseen data size, data dimension, or number of classes. 
\rh{We also demonstrate that our model can be adapted to coordinate ordering of other types of plots such as RadViz  by replacing the proposed shape-aware silhouette coefficient with the corresponding quality metric to guide network training. }

%% file: introduction.tex
\IEEEraisesectionheading{\section{Introduction}\label{sec:intro}}
\IEEEPARstart{S}{ince} the beginning of information visualization, radial plots have been used for showing multi-dimensional data points in complex data sets. Already in 1858, Florence Nightingale~\cite{Nightingale1858} applied them to depict mortality causes for the months of a year. 
\emph{Star glyphs} are a type of radial plot that show data points in a compact representation since the coordinate axes are circularly arranged around a central point. Variants such as RadViz~\cite{Hoffmann1997} and ArcVis~\cite{Long2018} also enable to visualize multiple dimensions in 2D radial plots.
An overview of radial visualization methods and glyph-based plots is given by Behrisch et al.~\cite{Behrisch2018} and Borgo et al.~\cite{borgo13}.

\od{Similarity search and grouping are among the most common analysis tasks for star glyph plots~\cite{andrienko2006exploratory,fuchs14,miller19}. Both tasks involve comparing several plots within a \emph{set} of glyphs. For example, the task may involve the search of glyphs with similar characteristics, or the separation or clustering of glyphs into different groups based on the data}. The order of axes in a glyph heavily influences its shape and thus its perception during such tasks. Thus, many methods have been proposed to optimize the ordering of axes~\cite{miller19,peng2004,klippel2009}. However, finding a suitable ordering poses some challenges. 
Ankerst et al.~\cite{ankerst1998} show that for different types of plots including parallel coordinates, finding an optimal ordering of axes according to a quality measure is an NP-hard problem, since it is equivalent to the Traveling Salesman Problem (TSP). Similar arguments apply to star glyphs and other radial plots. Thus, dimension ordering is a difficult combinatorial problem and heuristic algorithms are needed to solve it efficiently. However, most heuristic algorithms have been used to order the axes of \emph{individual} glyphs, without considering the global view of the entire \emph{set} of glyphs. For tasks such as similarity search or grouping within \emph{labeled} datasets, optimizing axes according to the entire set would allow to provide a better visual separation between the various classes. \od{In this scenario, the glyphs already possess different category labels, and the visual analysis task involves finding relationships between the data and the classes, in order to explain the grouping of data points.}

The desired shape of a star glyph also depends on the visualization task. While for a single star glyph, features like monotonicity, symmetry, and the presence of spikes are key criteria to consider when reordering axes~\cite{peng2004}, for similarity search and grouping among a set of star glyphs, especially when associated to multiple class labels, the perceptual differences between glyphs (their global class separation) becomes more important (see Fig.~\ref{fig:teaser}). Thus, in this paper we focus on a set 
of star glyphs instead of a single glyph, and aim to maximize visual class separation by optimizing the coordinate ordering of the glyphs. Thus, in contrast to existing works, we optimize the coordinate ordering of glyphs by considering an entire set, rather than just individual plots.





To measure the perceptual distance between two star glyphs in a set, we use shape context descriptors~\cite{belongie2001shape}.  These descriptors were developed in computer vision to measure the differences (or distances) between shapes to aid shape matching problems. 
To build shape context descriptors, first we sample a set of points on the boundary of each star glyph. Then, for each sampled point, we compute a 2D distribution of the relative positions of all remaining points. The average matching cost between each pair of corresponding sampled points on the two shapes is defined as the shape distance.

To measure the class separation of a set of star glyphs, we further employ a \textit{silhouette coefficient}~\cite{kaufman2009finding}, based on the shape distance given by shape context. 
However, computing the silhouette coefficient requires to compute shape context distances for each pair of shapes, which is time consuming. To make this computation more efficient, we train and use a neural network which estimates the shape context distance between any two given shapes.

With class separation as our objective function, we present a coordinate ordering method to maximize the silhouette coefficient. We introduce a set-to-sequence network inspired by 
Nazari et al.~\cite{nazari2018} to solve the combinatorial optimization problem. 
The network takes a set of data points with associated class information as input, and outputs the optimized coordinate order with a maximal value for the silhouette coefficient.
The network consists of an encoder and a decoder with an attention mechanism. The encoder employs a recurrent neural network (RNN) to encode the input shape and cluster information, while the decoder together with the attention mechanism employs another RNN to output a sequence with the coordinate ordering. 
The network is trained via reinforcement learning with the silhouette coefficient as the reward. 
In comparison to existing methods, which use exhaustive or random search to optimize  coordinate ordering, we believe that the advantage of a machine learning-driven approach is that it can learn a set of rules for a more efficient ordering, and our experiments seem to support this.

Since the method allows us to efficiently find good coordinate orders, it enables interactive optimization even for sets with a large numbers of dimensions such as 32.
We conducted two user studies to show that the silhouette coefficient based on shape context distance is a good measure for class separation, and that the coordinate orders provided by our method are preferred by users.
We tested our model on different settings to demonstrate its robustness and generalization abilities and show that it allows to order input sets with unseen data size, data dimension, or number of classes. \od{Our model also provides good coordinate ordering for other types of plots such as RadViz. In this case, the  class separation measure can be replaced with alternatives relevant to tasks involving this type of plot.}
In summary, the main contributions of this paper include: 
\begin{itemize}
    \item \change{The introduction of a set-to-sequence network trained via reinforcement learning for efficiently optimizing coordinate orderings of star glyphs to maximize class separation;}
	\item \change{The use of shape context descriptors for measuring the perceptual distance between two star glyphs and a silhouette coefficient for measuring the class separation of a set of star glyphs;}
        \item \change{The introduction of a neural network to more efficiently estimate the shape context distance between any two given shapes;}
	\item A user study to demonstrate that the proposed silhouette coefficient based on shape context distance is a good measure for class separation and
	 that the coordinate orders provided by our method are better than the ones obtained using baseline algorithms.
\end{itemize}

%% file: related.tex
\section{Related Work}
\label{sec:related_work}

\input{figures/overview}

A important line of work in visualization research has proposed methods for measuring the visual quality of data plots. These measures can then be used within optimization frameworks to improve the quality of the plots in different ways. For example, quality measures can guide the selection of an informative subspace when performing dimensionality reduction, or can provide an improved ordering of the axes of certain plots. Our work focuses specifically on the problem of improving axis ordering of sets of star glyphs and other radial plots, and thus we discuss work related to axis reordering in more detail. After that, we discuss the use of neural networks for solving combinatorial problems. 

\subsection{Axis reordering of data plots}

As mentioned above, Ankerst et al.~\cite{ankerst1998} showed that optimal ordering of axes is an NP-hard problem. They demonstrated  it for parallel coordinates, but similar arguments apply to star glyphs. Heuristic algorithms to solve this problem can be used for different types of visualizations, while most quality measures are tailored towards specific types of plots.

For {\em star glyphs}, Miller et al.~\cite{miller19} summarize four major strategies that are typically used for ordering axes: (1) User-driven criteria based on domain knowledge; (2) Correlation- and similarity-driven strategies that place similar axes close to each other; (3) Spike and salient shape strategies that place dissimilar axes close-by; and, (4) Symmetry-driven methods that build symmetric plots. Variations of the last three criteria can be computed automatically and optimized by heuristic algorithms. For example, Peng et al.~\cite{peng2004} reorder axes of star glyphs to reduce the amount of visual clutter using a strategy that seeks to improve monotonicity and symmetry of each glyph. Reordering is performed with a random swapping heuristic that starts with an initial order and randomly switches the positions of two dimensions. Klippel et al.~\cite{klippel2009} design glyphs with different non-convex salient shapes and test them for classification tasks, concluding that certain salient shapes are more helpful for classification than convex shapes, however, no specific algorithm is proposed here for axis reordering. 

Note that all of the approaches discussed above focus on the structure or shape characteristics of a single data point (single plot). In contrast, our approach compares the differences among a set of data points based on the shape context descriptor~\cite{belongie2001shape}. Thus, we reorder the dimensions of star glyphs according to a more global view. In addition we employ recently developed neural-network-based methods that allow to solve a variety of combinatorial problems, providing better results when compared to local, greedy heuristics. 

Regarding other types of plots, Ankerst et al.~\cite{ankerst1998} optimize the dimension ordering of parallel coordinates, circle segments, and recursive visualization patterns with an ant colony algorithm~\cite{dorigo1997ant}. They use Euclidean distance as the measure of similarity between two dimensions, so that their method places similar axes nearby. Artero et al.~\cite{artero2006} reorder axes of parallel coordinates and what they call Viz3D plots  based on axis similarity using a nearest neighbor heuristic. Axis similarity is measured either using the  L1-Norm or Pearson's Correlation Coefficient. Tatu et al.~\cite{tatu2009} reorder the axes of parallel coordinate plots with a measure based on the Hough Transform and the A* Algorithm. Albuquerque et al.~\cite{albuquerque2010} introduce a greedy incremental algorithm to sort the order of axes for Radviz. Their method starts by creating a Radviz with only two dimensions, adding the remaining dimensions one by one according to the guidance of a cluster density measure. Di Caro et al.~\cite{di2010} use the Davies-Bouldin Index to measure the quality of Radviz plots, where the index measures the inter- and intra-cluster separation. Dasgupta and Kosara~\cite{dasgupta2010} propose a set of screen-space metrics called Pragnostics (parallel coordinates diagnostics) to quantify visual structures that a user may perceive in parallel coordinate plots, such as line crossings or parallelism of lines. Their method allows to optimize  plots according to different user preferences. Note that none of the methods discussed here is suitable for coordinate reordering of star glyphs, since the quality measures employed do not measure the perceptual difference between two plots, but rather measure other properties of the plots such as axis similarity or clustering measures.

\subsection{Neural nets for solving combinatorial problems}
Pointer Networks (Ptr-Net)~\cite{vinyals2015} are one of the best-known neural architectures for solving combinatorial problems. They form supervised models that can compute convex hulls, Delaunay triangulations, and TSP solutions. However, given the difficulties in data acquisition, supervised learning is not the best way of solving combinatorial optimization problems. Thus, Bello et al.~\cite{bello2017} incorporate reinforcement learning into Ptr-Nets to yield an unsupervised learning paradigm. Ptr-Net and its variations use an RNN encoder to receive a sequence of elements as input and then output another sequence (such as a reordering of the elements). For classic combinatorial problems, most inputs are sets of data points without any ordering information. Thus, Nazari et al.~\cite{nazari2018} argue that the RNN encoder adds an extra complication to the encoder and is not really necessary. They remove the RNN encoder and propose a set-to-sequence network. Such a set-to-sequence network is suitable to solve our dimension ordering problem and thus we base our approach on this architecture.

%% file: figures/overview.tex
\begin{figure*}[tb]
    \centering
    \includegraphics[width=0.99\linewidth]{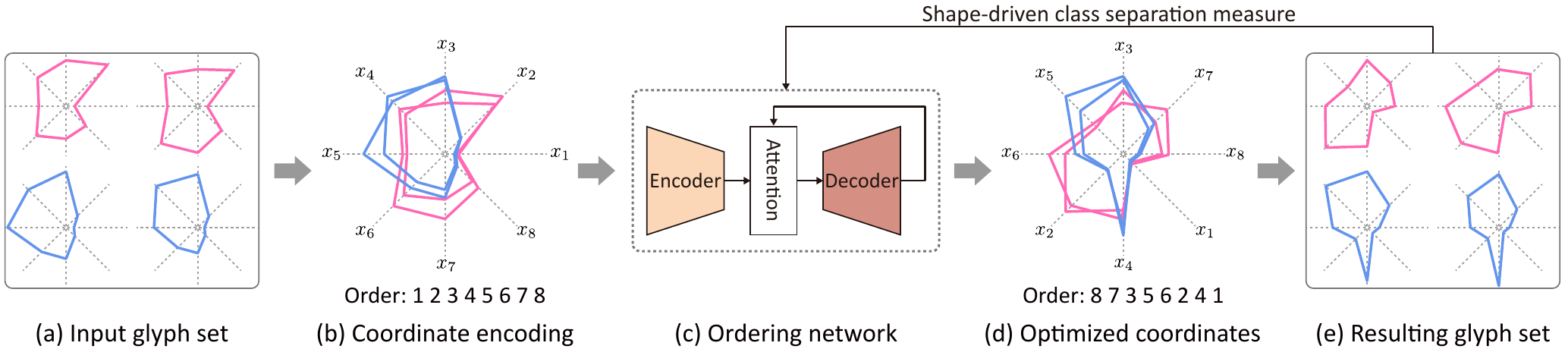}
	\caption{\od{Overview of our method, illustrated using a data set of four 8D data points with 2 class labels indicated by red/blue colors. 
    Given a set of data points  with associated coordinate ordering (a), we first accumulate the data information along each coordinate (b), and then feed this information
    in sequences  to the ordering network (c) to get an optimized coordinate order (d).  The optimization of the network is guided by the shape-driven class separation measure, i.e., the silhouette coefficient based on the shape distance given by the shape context, computed on the resulting set of star glyphs with the new coordinate order (e).}}
	\label{fig:overview}
\end{figure*}

%% file: method.tex
\section{Method}
\label{sec:method}

\od{As mentioned above, our goal is to optimize the coordinate order so that class separation is maximized when visually inspecting a set of glyphs, i.e., intra-class shape distances should be minimized while inter-class distances maximized. Thus, a shape-aware class separation measure is defined to guide the optimization. The details of the shape distance measure are given in Sect.~\ref{sec:sc}, while the derived class separation measure is introduced in Sect.~\ref{sec:cs}.

Using the shape-aware class separation measure as our objective function, we introduce an ordering network 
using reinforcement learning. 
To find the optimized coordinate order for the input data set, we first 
accumulate
the values along each coordinate and then feed this information in sequences into the ordering network. 
Fig.~\ref{fig:overview} gives an overview of our learning-based ordering pipeline. More details about the coordinate encoding and network structure are found in Sect.~\ref{sec:architecture}. The training details are given in Sect.~\ref{sec:train}. }




\subsection{Shape distance measure}
\label{sec:sc}

When visualizing a multi-dimensional data point using a star glyph, its shape can be seen as an abstract representation of the data. To measure the differences between two shapes, we adopt a shape descriptor from computer vision and geometric processing called shape context \cite{belongie2001shape}, which is commonly used for shape matching.

\input{figures/sc}

In more detail, given a star glyph, we first abstract it into a shape $S$, represented as a 2D polygon. Next, we uniformly sample $h$ points on the boundary of $S$. For each sampled point $p_t$, we compute the positions of the remaining $h-1$ sampled points relative to $p_t$ and compute a 2D distribution of these points (see Fig.~\ref{fig:sc}).  
Specifically, we create a log-polar histogram of the relative positions by encoding the points in polar coordinates and counting the number of points falling into each bin of the histogram. For a working histogram, we chose 5 bins for the logarithm of the radius $r$ and 12 bins for the angle $\theta$. The maximum radius stored in the histogram is two times the average distance between point pairs.
To compute the distance between two shapes, we employ the $\chi^2$ test, a statistical distribution matching function, between the histograms of the two shapes. The distance can be regarded as a ``shape context cost'' for matching two data points:
\begin{equation}
\label{eq:Cost}
\text{Cost}(\text{hist}_1, \text{hist}_2) = \frac{1}{2}\sum_{b=1}^{B}{\frac{\left({\text{hist}_1(b) - \text{hist}_2(b)}\right)^2}{\text{hist}_1(b) + \text{hist}_2(b)}},
\end{equation}
where $\text{hist}_1$ and  $\text{hist}_2$ are the normalized shape context descriptors of two points, and $B$ is the total number of bins used in the shape context descriptor. In our work we set $B = 5 \times 12 = 60$, as discussed above. The values of the cost are normalized to a range from 0 to 1. Next we compute the average cost for all the $h$ pairs of points sampled over the boundaries of the two glyphs as the shape distance, where we assume that the points are in correspondence along the boundaries:
\begin{equation}
\label{eq:scd}
d(S_i, S_j) =\frac{1}{h}\sum_{t=1}^{h}{\text{Cost}(\text{hist}(p^i_t), \text{hist}(p^j_t))},
\end{equation}
where $\{p^i_t\} _{t=1}^{h}$ is the set of points sampled on shape $S_i$, and $\text{hist}(p^i_t)$ is the normalized shape context descriptor of point $p^i_t$. Note that in contrast to the original definition, we do not make the descriptor invariant to rotations, since star glyphs have a natural orientation and are not read from different angles.

\subsection{Class separation measure}
\label{sec:cs}
For a set of shapes with different class labels, we first define a \emph{silhouette value $s_i$} to measure how similar shape $S_i$ is to its own cluster (cohesion) compared to other clusters (separation):
\begin{equation}
s_i = \frac{b_i - a_i}{\max{\{a_i, b_i\}}},
\label{eq:silhouettevalue}
\end{equation}
where $a_i$ is the mean distance between $S_i$ and all the other data points with the same class label, and $b_i$ is the smallest mean distance of $S_i$ to all other class clusters: 
\begin{equation}
a_i = \frac{1}{|C_i| - 1}\sum_{j \in C_i, j \neq i}{d(S_i, S_j)},
\end{equation}
\begin{equation}
b_i = \min_{k \neq i}{\frac{1}{|C_k|}\sum_{j \in C_k}{d(S_i, S_j)}},
\end{equation}
where $C_i$ indicates the cluster $S_i$ belongs to, and $d(S_i, S_j)$ is the shape distance measure defined in Eq.(\ref{eq:scd}).

We use the silhouette coefficient $SC$~\cite{rousseeuw1987silhouettes, kaufman2009finding} to measure class separation, which is defined as the maximum value of the mean silhouette value over all points of the entire dataset:
\begin{equation}
\label{eq:SC}
SC = \max_{k=1}^{K}\hat{s}_k,
\end{equation}
where $\hat{s}_k$ represents the mean silhouette value ($s_i$ from Eq. \eqref{eq:silhouettevalue}) over all data points with class label $k$, and $K$ stands for the number of different clusters. Note that $SC$ is positive by the definition. We use the silhouette coefficient $SC$ as our class separation measure since it is commonly used for this task, but any other class separation metric can be used to replace $SC$. 


\subsection{Coordinate ordering network}
\label{sec:architecture}

\od{
A set of high-dimensional data points from different classes can be represented by a matrix $P \in \mathbb{R}^{m \times n}$ associated with class labels $L \in \mathbb{R}^{m}$, where $m$ is the number of multi-dimensional data points (or star glyphs), while $n$ is the data dimensionality (number of coordinates) that we need to order. $L_i \in \{1,2,\dots,K\}$ is the class label of the $i$-th star glyph and $K$ is the number of different classes. 

To put more focus on the coordinates we are about to optimize, we re-encode the input information coordinate-wise instead of point-wise. The coordinate encoding is denoted by $X = \{x_i\}$, where $x_i = [p_i, c_i]\in \mathbb{R}^{m \times 2}$ represents the $i$-th coordinate value $p_i$ for $m$ input points and their corresponding cluster information $c_i$. Note that $p_i $ is the $i$-th column of the input data matrix $P$. However, we decided not to use the associated class label matrix $L$ as $c_i$, since we only need the cluster information determined by the class labels rather than the labels themselves. Thus, instead of using $L$ directly, we compute the cluster centers for each of the $K$ classes in the $n$-dimensional input space, and then associate each point with the center of its cluster. This results in a matrix $C \in  \mathbb{R}^{m \times n}$ with $c_i $ being the $i$-th column of $C$. Fig.~\ref{fig:input} illustrates the input encoding.
}
\input{figures/input}

\input{figures/network}

\od{Our goal is to find a stochastic method that generates the coordinate order in a way that maximizes  class separation. Thus, we adopted a neural optimization model for this purpose that is trained with reinforcement learning.}
Fig.~\ref{fig:network} illustrates the architecture of our network for the optimization of the coordinate order. The network takes as input the coordinate encoding of a set of multi-dimensional data points with their cluster-membership information, and outputs an optimized sequence of the coordinates. Similar to the approach by Nazari et al.~\cite{nazari2018}, our network consists of an encoder and a decoder with an attention mechanism linking the two modules. The encoder maps each $x_i$ into a higher-dimensional space and passes it to the attention mechanism. The decoder outputs an optimized sequence of coordinates $Y = \{ y_i \}$, where $y_i$ is the index of the coordinate selected for position $i$ in the optimized sequence.

Please note that the network originally designed by Nazari et al.~\cite{nazari2018} allows to take data points with different dimensions $n$ as input. However, since the size $m$ of the input sets changes with different data sets, we need to additionally enable our network to take $X$ with various sizes. To achieve this, we use RNNs in the encoder as well as the decoder.


To compute an optimized sequence $Y$ from our input $X$, the network iterates $n$ times over the input. For each iteration $t$, the attention mechanism outputs a probability map over all coordinates, which indicates the probability that each coordinate is the best choice for entering the sequence at this point. The attention mechanism determines the probabilities from the input $X$ as well as information that was accumulated internally until iteration $t-1$. The coordinate with the maximum probability is then selected to be the output $y^t$ at iteration $t$ and eliminated from the set of coordinates that 
will be further considered by the attention mechanism. Fig.~\ref{fig:network} shows an example of the input processing. The first coordinate $x_1$ is selected as output $y^t$ by the attention mechanism (high probability indicated by the red bar), and thus $x_1$ is no longer available for selection in the next step. In order to select each coordinate only once, a masking scheme is used, which sets the logarithmic probabilities of invalid states (already taken coordinates) to $-\infty$. The process iterates until a sequence $Y$ of $n$ coordinates is obtained.

\subsection{Network training}
\label{sec:train}

\od{Following the standard nomenclature of reinforcement learning, in our method, the \emph{actions} are the coordinate selections.}
\od{The \emph{state} is defined as the decoder output at each time step, where the decoder consumes the network input and the previously selected subsequence of coordinates. The \emph{reward function} is the silhouette coefficient $SC$ defined in Eq.~\eqref{eq:SC}, which measures the class separation quality of the resulting star glyphs with new coordinate order and is computed  only after the full sequence of actions was sampled.}
To train the network, we use the well-known policy gradient approaches as demonstrated by Nazari et al.~\cite{nazari2018}. This training method is standard in reinforcement learning, so we provide the details in the supplementary material and only discuss the ordering reward here. 


\input{figures/prediction}

Computing the $SC$ for the reward function requires to compute shape context distances for each pair of shapes, which is extremely time consuming, especially when the network is trained on a large-scale data set. To enable a more efficient training, we pre-train a second network to estimate the shape context distance between any two given shapes. The network is similar to the encoder in Fig.~\ref{fig:network}, but connected to two fully-connected layers and a sigmoid layer; see Fig.~\ref{fig:prediction}. The sigmoid layer predicts a single value for the shape context distance. This network is trained with the mean-squared error (MSE) as the loss function.

To make the prediction more robust and be able to predict the distance for shapes with different numbers of dimensions $n$, we take the sampled points as input (as discussed in Sect.~\ref{sec:sc}) instead of the original corner points, although theoretically the RNN module would be able to take inputs with varying sizes. In Sect.~\ref{sec:sc_result}, we show that taking sampled points instead of corner points yields better ordering results.

%% file: figures/sc.tex
\begin{figure}[tb]
    \centering
    \includegraphics[width=0.99\linewidth]{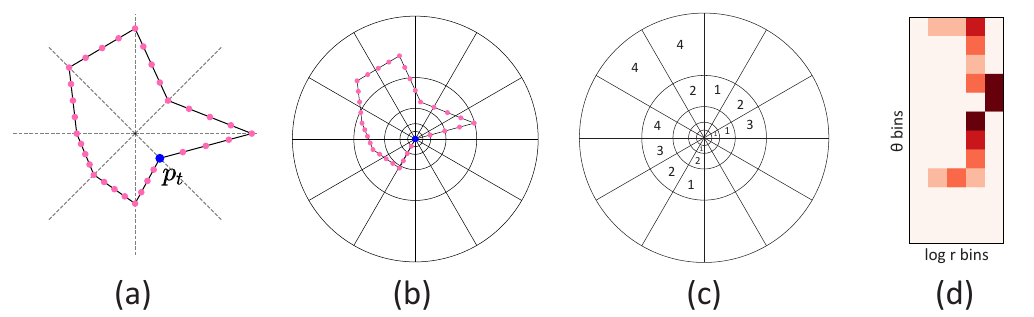}
	\caption{Shape context computation: (a) We sample the contour of the star glyph; (b) For each sample point $p_t$, we compute the relative position of all the other sample points; (c) We count the number of points falling into each bin of a log-polar grid centered at $p_t$, to produce a spatial histogram; (d) The histogram is stored as a two-dimensional matrix.}
	\label{fig:sc}
\end{figure}

%% file: figures/input.tex
\begin{figure}[tb]
    \centering
    \includegraphics[width=0.98\linewidth]{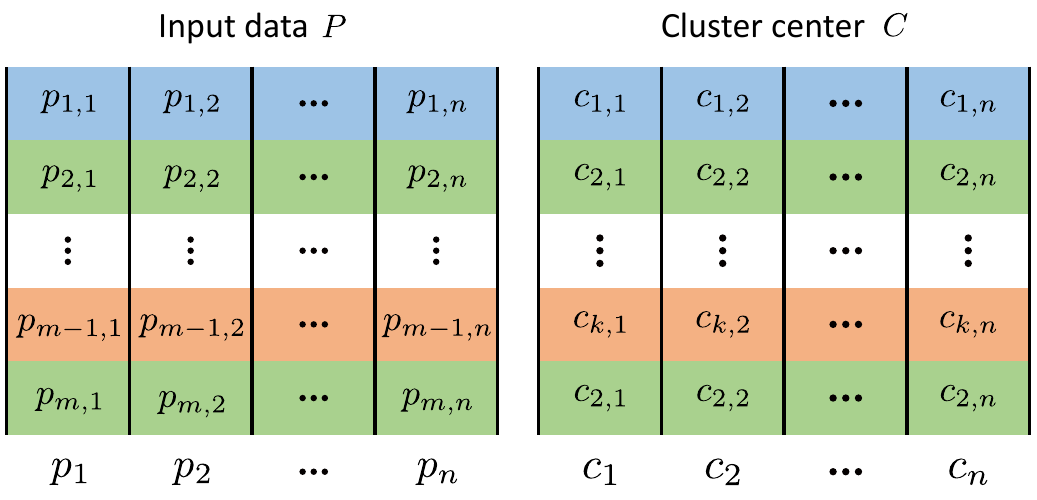}
    \caption{Input encoding along coordinates. The matrix $P$ captures a set of $m$  points with $n$ dimensions, while $C$ captures the $n$-dimensional center of the cluster associated to each point. Rows with the same color indicate points of the same class, which share the same cluster center.}
    \label{fig:input}
\end{figure}

%% file: figures/network.tex
\begin{figure}[tb]
    \centering
    \includegraphics[width=0.98\linewidth]{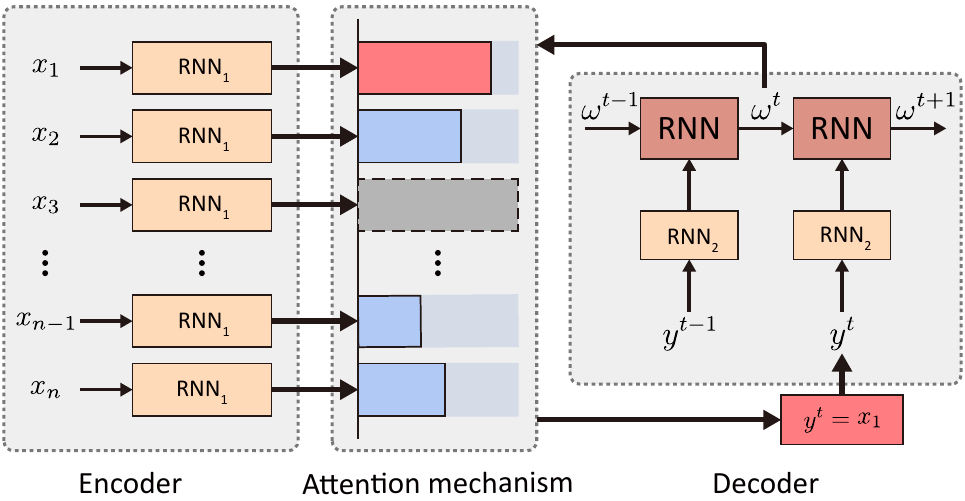}
    \caption{Architecture of our neural network for coordinate order optimization. The network takes as input a set of point coordinates and cluster assignments $\{ x_i \}$, and outputs an optimized coordinate sequence $\{y_i\}$. Please refer to the text for details on the components of the architecture.}
    \label{fig:network}
\end{figure}

%% file: figures/prediction.tex
\begin{figure}[tb]
    \centering
    \includegraphics[width=0.98\linewidth]{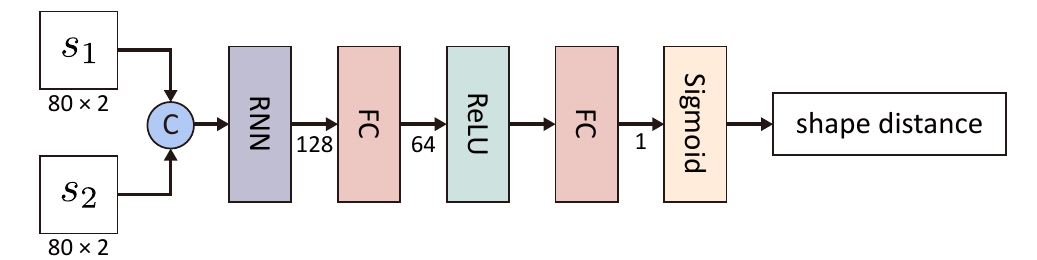}
	\caption{Shape context distance prediction network, which estimates the scalar distance between two shape context descriptors $s_1$ and $s_2$, where ``C'' denotes concatenation and ``FC'' is a fully-connected network.}
	\label{fig:prediction}
\end{figure}

%% file: results.tex
%

%
\section{Silhouette Coefficient: Results and Evaluation }
\label{sec:sc_result}


\subsection{Approximation of the shape context distance}
The shape context distance is used to compute the reward function that guides coordinate optimization. Thus, we first have to test the approximation accuracy of the distance approximation network described in Section~\ref{sec:train}. 

To generate the star glyphs used for training the network, we produced multivariate data by randomly sampling the dimension $n$ from the range $[16,24]$ and then randomly sampling the value of each dimension from $[0,1]$. To compute the ground-truth shape context distance between any pair of training glyphs, we sampled $h=80$ points on each shape contour and then computed the shape context distances as discussed in Section~\ref{sec:sc}. In total, we generated 64k shape pairs for training, requiring 0.6 GB of memory. The training time was 10 min, while the final size of the network in memory is 0.3 MB.

To justify the use of uniformly-sampled points as input to the network, we also trained our network with the same set of training data but computed the shape context distances using the original corner points, i.e., dimension values. We compared the training results of these two networks, where the corresponding settings are denoted as SAMPLE and ORIG.

\input{figures/sc_comp}

To create the test sets, we generated shape pairs in the same manner as described above but with $n$ values sampled from a larger range $[16,32]$, to evaluate whether the prediction network is able to generalize well for shapes with different numbers of input dimensions. For each $n$, we generated 10k shape pairs. Then, we tested the networks trained with the two different representations of input data and computed the average MSE on the test data to measure the prediction accuracy. 
Fig.~\ref{fig:sc_comp} shows the comparison. We see that the SAMPLE setting consistently provides low errors for various input sizes, while ORIG only works well for inputs with dimensions within or close to the training range.

To evaluate the gain in efficiency by using the distance approximation network, we compared the running time of shape distance computation with and without using the approximation network on one CPU. For a pair of star glyphs, the average running time was 0.21 ms using the network and 11.69 ms without, showing that the network is more than $50$ times faster than the direct distance computation. Note that computing the silhouette coefficient $SC$ for a set of star glyphs requires to compute shape context distances for each pair of shapes, which is extremely time consuming, especially when the coordinate ordering network is trained on a large-scale dataset with $SC$ as the reward function. \od{Thus, this approximation network is necessary for training the coordinate ordering network.}

\subsection{User study I: Silhouette coefficient}
We conducted a user study to justify using the silhouette coefficient $SC$ based on shape context distances as a class separation measure.

\paragraph{Datasets.}
We randomly synthesized 10 datasets with various sizes, dimensions, and class numbers. We randomly sampled the variables as follows: dimension $n \in [8,32]$, class number $K \in [2,4]$, and set size $m \in K \times [3, 6]$.  For producing synthetic datasets, we sampled all the data associated to the same label from the same Gaussian distribution, with a mean $\mu$ uniformly sampled from $[10, 100]^n$ and standard deviation $\sigma$ sampled from $[0.1, 0.3]$. In all the experiments below that use synthetic data, we generate the data in the same way.

\paragraph{Setting.}
Similarly to the setting of Miller et al.~\cite{miller19}  we plot the data using star glyphs with different coordinate orders for each set and ask the users to manually group the glyphs into clusters, where the number of clusters is also decided by the users. 

Since the user study of Miller et al.~\cite{miller19} shows that users perform better when glyphs have salient shapes with spikes that are obtained by a dissimilarity-based ordering of the dimensions, we generate one coordinate ordering based on the dissimilarity of the dimensions (marked as \emph{salient}). 
%
We ran an exhaustive search to find the coordinate ordering with the highest average dissimilarity between adjacent coordinates, where the dissimilarity is measured using Euclidean distance.
Moreover, we also generated two additional coordinate orderings with our method, according to high and low silhouette coefficients $SC$s (denoted as \emph{High SC} and \emph{Low SC}, respectively). 
This setup resulted in 3 different plots for each dataset and 30 sets of star glyphs in total. We asked 20 participants to participate in this user study, all of which were either undergraduate or graduate students with little or no experience in Information Visualization.

\paragraph{Results.}
We use the \textit{quality of groups} as the quality measure to assess the user performance with each coordinate ordering, following Miller et al.~\cite{miller19}. The computation of the quality of groups is based on the Jaccard index between the user grouping and the ground-truth grouping. Here we compute the average Jaccard index from two directions: the index of each group to its best match in the ground-truth and the  index of every ground-truth cluster to its best match in the selection. 
The final quality measure is the average score of these two Jaccard indices~\cite{miller19}.

\input{figures/user_study}

Fig.~\ref{fig:user_study} shows the statistics of the grouping quality computed on the results. The  quality of the sets with high $SC$s is 0.757, while the low-coefficient ones only achieve a quality of 0.697; the salient glyphs achieve 0.675. 
Note that, since the datasets used in this study have various sizes, dimensions, and class numbers, the difficulty levels for the grouping task for different datasets can vary significantly, leading to a noticeable high variance in the results.

In general, we find that the \emph{Salient} ordering 
only works well for grouping low-dimensional data where spikes are clearly noticeable. With increasing dimensionality of the data, user performance with the \emph{Salient} ordering decreases. \rh{On the other hand, our method provides more stable results for a variety of datasets and gets the best user performances in 80\% of the cases.


\input{figures/failure_case}

One case where a \emph{Salient} ordering enables users to achieve a higher grouping quality than \emph{High SC} is shown in Fig.~\ref{fig:failure}. It involves a dataset with dimension $n=16$, class number $K=4$, and set size $m=20$. 
To explain this result, we checked the mean silhouette value for each class and found that Class 3 has a low value while all the other classes have relatively high mean silhouette values. Our SC score is defined as the maximum value of the mean silhouette values of all clusters. 
Thus, the high values of three classes dominate the mean and therefore the low value of Class 3 does not stand out. This explains why a higher SC did not lead to better results here. Improving the results for these types of cases could be one direction to explore in future work.

In conclusion, our results show that the silhouette coefficient $SC$ based on shape context distances is a good and stable measure for class separation. In most situations, users were able to tell the perceptual difference between different glyphs more clearly for sets ordered with a high $SC$.}

\section{Coordinate Ordering: Results and Evaluation}

\od{We evaluated the general effectiveness of our coordinate order optimization with a comparison to a state-of-the-art baseline method and a user study. 
Then, we evaluated different aspects of our method in more depth.
The following experiments show that our method generates better results in a more efficient way and generalizes well to handle larger datasets.}

\input{figures/tab_comp}

\input{figures/comp_baseline}

\input{figures/baseline}

\paragraph{Datasets.}
Our method is trained using synthetic datasets and evaluated on both synthetic and real datasets. For producing synthetic datasets, we generate a collection of sets of multivariate data $P \in \mathbb{R}^{m \times n}$ associated with class label $L \in \mathbb{R}^{m}$ ($L_i \in \{1,2,\dots,K\}$), where the set size $m$, data dimension $n$, and the class number $K$ all vary across different sets. 
In all the experiments below using synthetic data, we trained the network using 64k sets and test on separate 10k sets. We use the Sensorless Drive Diagnosis Data Set~\cite{Dua:2019} as an example of real data, which has 58,509 49-dimensional data points from 11 classes. We randomly sampled 50 subsets with set size $m=8$, data dimension $n=16$, and class number $K=2$ to constitute the test set. Training time was 3.6 hours and the final size of the network in memory is 4 MB.

\subsection{Comparison to baseline} 
We compare our method to a state-of-the-art baseline method, the random swapping algorithm used by Peng et al.~\cite{peng2004clutter}, to show that our learning-based approach yields a better ordering than methods based on heuristic optimization. In short, starting with an initial order, the baseline algorithm randomly chooses two dimensions to switch. If the new order obtained by switching the two dimensions has a higher $SC$, then it is kept, otherwise the old order is left intact and the algorithm attempts to swap another pair. This process is iterated until no better result can be generated for a certain number of swaps (which we set as 10) or when a maximum number of iterations is reached (which we set as 100). 
We selected these parameter values to enable a reasonable execution time for the baseline.
We compare our method to this baseline on synthetic and real datasets by using $m=8$, $n=16$, and $K=2$. Note that the network is trained only using synthetic data, but is tested on both types of  datasets.

The results are shown in Table~\ref{tab:comp}. For synthetic data, the $SC$ of our method is 0.672 and the $SC$ of the baseline is 0.641, while for the real data, the $SC$s are 0.478 for our method and 0.461 for the baseline. Moreover, once the network is trained, applying it to a test glyph takes 0.02s, while it takes on average 2.93s to apply the random swapping baseline algorithm.  Fig.~\ref{fig:comp_baseline} shows a visual comparison of our method with the baseline on a few examples of synthetic and real datasets. Our results have higher $SC$s and the class separation is more prominent. The example on the bottom is challenging since the values of some coordinates are similar for data from different classes. Nevertheless, our method is able to group the coordinates with more dissimilar values together, so that the overall difference among the shapes is clearer, producing peanut-shaped and fan-shaped glyphs in this case.


Note that we set the parameters of the baseline method so that the results can be obtained in a few seconds. If we select larger values for the parameters, the results tend to improve but take longer time. Fig.~\ref{fig:baseline} shows how the performance of the baseline method changes with the increase of the maximal number of swap trials $n_s$ in each iteration, where the maximum $n_s$ is set to 1,000. The figure shows the average of $1,000$ experiments for each $n_s$.
For reference, we ran an exhaustive search to obtain the ground-truth, which on average has a $SC$ of 0.681. The average $SC$ of our method is 0.672, only slightly below this optimal value. The performance of the baseline method increases with larger $n_s$ and reaches similar performance to our method when $n_s = 55$. However, the method needs more than 14 seconds to obtain such a result (compared to 0.02s of our method).

\subsection{User study II: Optimized coordinate orderings}
\change{We conducted another user study to demonstrate that our sets of star glyphs with optimized coordinate order allow to analyze data more effectively, and show that the visualization of our star glyphs is preferred over the original ordering and the one obtained using the baseline~\cite{peng2004clutter}. }

\rh{We randomly selected 30 datasets with various sizes, dimensions, and class numbers, as in User Study I, but with half synthetic and half real data. For synthetic data, the original order is random.
Each dataset was first presented in two versions to the users (in random order): one with the original coordinate order and one with coordinate order optimized by our method. 
The users were asked to choose the visualization that they thought better identifies shapes from different classes. Note that star glyphs with different classes were shown in different colors  and we did not provide any other instructions. Users could select either one of the visualizations or ``Cannot decide''.}

We asked 30 participants to do this user study, all graduate students in computer science. A total of 900 answers were collected.
The votes for the options ``our/original/cannot decide'' were 661/172/67, respectively.
Thus, for 73\% of the questions, the participants thought that the visualization with optimized coordinate ordering was better than the initial ordering, which confirms the effectiveness of our method. 

\change{To further compare with the baseline method~\cite{peng2004clutter}, we also asked  users to select between the results obtained by our method and the baseline for all the 30 datasets. The votes for the options ``our/baseline/cannot decide'' were 546/294/60, respectively.  Thus, for 61\% of the questions, the participants thought that the visualization obtained with our method was better than the baseline, while only 33\% preferred the baseline. This confirms that our method outperforms the baseline. After further analyzing the results, we found that the selection of datasets where the baseline's visualization was preferred was not coherent across participants.}


\subsection{Detailed evaluation of the network}
\rh{We evaluated different aspects of our coordinate optimization method, including the cluster representation and generality of the method to a variety of datasets. Experiments show that our choice of cluster representation provides better results and our model has a high generality to different data settings.}

\textit{Representation of cluster information.}
The cluster information given by the class labels can be represented in two different ways in the input given to the network. We can represent the clusters directly with their class labels, or we can represent the clusters with cluster centers, which are invariant to the class labels. We justify the use of cluster centers with the results of an experiment where we fixed $m=8$, $n=16$, and $K=2$, and compared the performance of the network using the two representations. We found that the coordinate ordering was slightly better when using the cluster centers ($SC=0.672$) than using labels directly ($SC=0.670$), where the $SC$ for the input order is 0.564.


%


\input{figures/diff_n}

\input{figures/star_result_n}

\input{figures/star_result_m}

\input{figures/diff_m}

\textit{Diversity of data sets.}
To test the generality of our method with regard to datasets of different size $m$, data dimension $n$, and class number $K$, we conducted experiments where we evaluated the change in performance of our method when varying one variable and keeping the other two fixed.

\textit{Experiments with different data dimension $n$}. We fixed the set size $m=8$ and class number $K=2$, and assigned 4 data points to label 1 and 4 data points to label 2. Then, we randomly sampled the data dimension $n$ from the range $[16, 24]$ to train the network (which we denote as $n=16\sim24$). 
Fig.~\ref{fig:diff_n} shows the performance for sets with different $n$. With our order optimization, the $SC$ of the optimized order is much higher than that of the original random order.  
We also compared to networks trained with a fixed dimension data of $n=16$ or $n=24$.  
\rh{The network trained on $n=16$ works best when testing on sets with the same dimensionality since in this case the testing set has the most similar distribution to the training set, but the performance drops quickly when $n$ increases as the differences between the test and training set become larger. Similarly, the network trained on $n=24$ provides the best performance when $n=24$, while the performance drops when either $n$ increases or decreases. When trained on $n=16\sim24$, the overall performance on sets with $n=16\sim24$ is best, and the performance drops when $n$ increases. However, the method still provides a performance comparable or even better than when trained with $n=24$. This indicates that the network trained with datasets that cover a larger distribution is able to generalize better. }

Fig.~\ref{fig:star_result_n} shows the comparison between results obtained using networks trained with different data dimension $n$. With the random input order, the difference between shapes from different classes is subtle, especially when the data dimension $n$ increases. However, after optimizing the coordinate order of the glyphs with our method, the class difference becomes much more prominent. When comparing results obtained from networks trained with different data, we see that the network trained on $n=16\sim24$ consistently provides good results for data sets with different $n$.

\input{figures/diff_k}
\input{figures/star_result_k}

\textit{Experiments with different set size $m$}. Here, we fixed the data dimension $n=16$ and class number $K=2$, and randomly sampled the set size $m$ from $[8, 16]$, with half of the data points assigned to label 1 and the other half assigned to label 2. 
We also compared to the setting where we trained the network with datasets having a  fixed size of $m=8$ or $m=16$.
Fig.~\ref{fig:diff_m} shows the comparison among all the networks. We notice that the performances of networks trained on $m=16$ and $m=8\sim16$ are similar and better than that of $m=8$, while $m=8\sim16$ consistently provides the best results. 
\rh{Moreover, both networks trained on $m=16$ and $m=8\sim16$ have consistent performances even when $m$ increases to 32. The performance of the network trained on $m=8$ drops only a bit, which means that different set sizes have less significant influence on network performance.}
Nevertheless, all dimension orders provided by networks trained with different $m$ are better than the original order. 

Fig.~\ref{fig:star_result_m} shows example results for networks trained with different $m$. Similar to the results shown in Fig.~\ref{fig:star_result_n}, with random order, it is difficult to tell apart the shapes from different classes. However, with an optimized order, the separation between different classes and the similarity within the classes are much clearer.

\textit{Experiment with different class number $K$}. Here, we fixed the data dimension $n=16$ and set size $m=32$, and randomly sampled the class number $K$ from $[2,4]$. We evenly distributed the points into $K$ classes. 
We also compared to the setting where the network was trained with a fixed number of classes $K=2$ or $K=4$.
Fig.~\ref{fig:diff_k} shows the comparison. When testing on data with $K=2$, the performance difference is clear: the network trained with $K=2$ is better than $K=2\sim4$, while $K=4$ provides the worst result. However, as the number of classes in the test set increases, the performance of the networks trained with $K=2\sim4$ and $K=4$ is close and better than the performance of the network trained with $K=2$.
\rh{Compared to $n$ and $m$, the change of $K$ leads to a more significant drop of the $SC$, which means that the change of the class number decreases the generality of our networks. 
Thus, it is better to train the network for different class numbers separately, which is reasonable since, in most of the settings, the class number tends to be fixed while the data size or dimension is more likely to vary.}

Fig.~\ref{fig:star_result_k} shows results obtained using networks trained with different $K$. 
When the number of classes is small, although there are some differences in the $SC$s, it is still relatively easy to see the difference between shapes of different classes for results obtained using different networks. However, when the number of classes increases, the results obtained with the network trained with either $K=2$ or $K=2\sim4$ show clearer differences, especially for the shapes of the first two classes.

\input{figures/radviz}
\input{figures/comp_radviz}

\input{figures/tab_comp_radviz}

\section{Generalization to RadViz}

Star glyphs are used to compare different data points in a set. However, if we would like to have a more global view of the distribution of a given dataset, RadViz plots can be used. RadViz is a multivariate data visualization algorithm that assigns each feature dimension uniformly around the circumference of a circle and then plots points on the interior of the circle such that each point normalizes its values on the axes from the center to each arc. This mechanism allows to plot as many dimensions as fit on a circle which greatly expands the possible dimensionality of the visualization. However, the dimension ordering problem also exists for these plots. To demonstrate the generality of our method, we therefore applied our network-based optimization  to RadViz plots. 

\rh{
To guide our method for ordering RadViz axes, we take the objective function used by Di Caro et al.~\cite{di2010analyzing} as the reward function, which is defined as the ratio between the Davies-Bouldin index~\cite{davies1979cluster} of the original data and for the 2D mapping.
As for star glyphs, we re-encoded every data set coordinate-wise and fed it into our network, which then outputs the optimized coordinate order. We only change the objective function that is used to evaluate the resulting plot  and guide the network training.}

We train the network with 25k datasets with set size $m=100$, data dimension $n=16$, and class number $K=4$, and test it on another set with 10k datasets. Both training and test sets were synthetically generated. The training time here is 7.2 hours and the final size of the network in memory is 4 MB. Fig.~\ref{fig:radviz} shows visual examples of coordinate orderings with our method on the synthetic data. We see that the separation between classes is much clearer with the optimized coordinate order.

Using real data sets, we compared our approach to the baseline method of random swapping. We used the Mice Protein Expression Data Set~\cite{Dua:2019}, where $m=824$, $n = 1,080$, and $K=8$. We randomly sampled 50 subsets of data with $m=100$, $n=16$ and $K=4$. The comparisons are shown in Table~\ref{tab:comp_radviz}. 
Our method consistently provides better results than the baseline method. Moreover, once the network is trained, applying  it takes only 0.07s  in comparison to 0.4s to apply random swapping. Our method is slightly slower when applied to RadViz (compared to star glyphs) since the input size $m$  of the training sets is larger, while the baseline is faster here, because it is faster to compute  the Davies-Bouldin index than the $SC$.
Visual comparisons of the two methods can be found in Fig.~\ref{fig:comp_radviz}.
The clusters are much better separated using our network than for the order obtained with the baseline method.

Note that the silhouette coefficient $SC$ can also serve as a cluster separation measure for RadViz. However, the Davies-Bouldin index cannot be used for star glyphs since the centroid of each cluster needs to be computed, while it is not clear how to define the centroid for a set of star glyphs. We also used the $SC$ to optimize RadViz plots and obtained similar results to those obtained with the Davies-Bouldin index.

%% file: figures/sc_comp.tex
\begin{figure}[tb]
    \centering
    \includegraphics[width=0.96\linewidth]{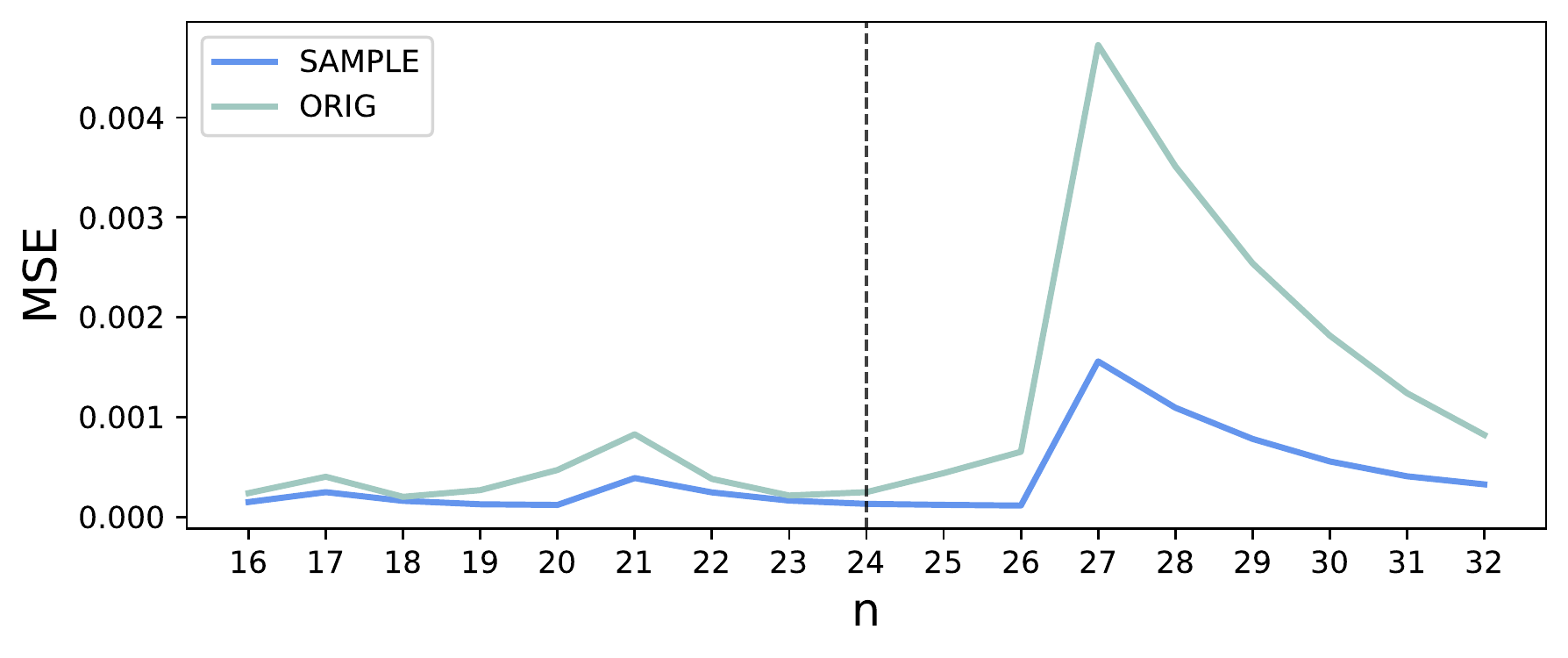}
	\caption{Comparison of shape context distance prediction results using two different representations of the shapes. ORIG indicates that the original corner points of the glyphs are used, where the dimension can vary for different data sets,  while SAMPLE indicates that a consistent number of points are sampled from the shape contour.}
	\label{fig:sc_comp}
\end{figure}

%% file: figures/user_study.tex

\begin{figure}[tb]
    \centering
    \includegraphics[width=0.9\linewidth]{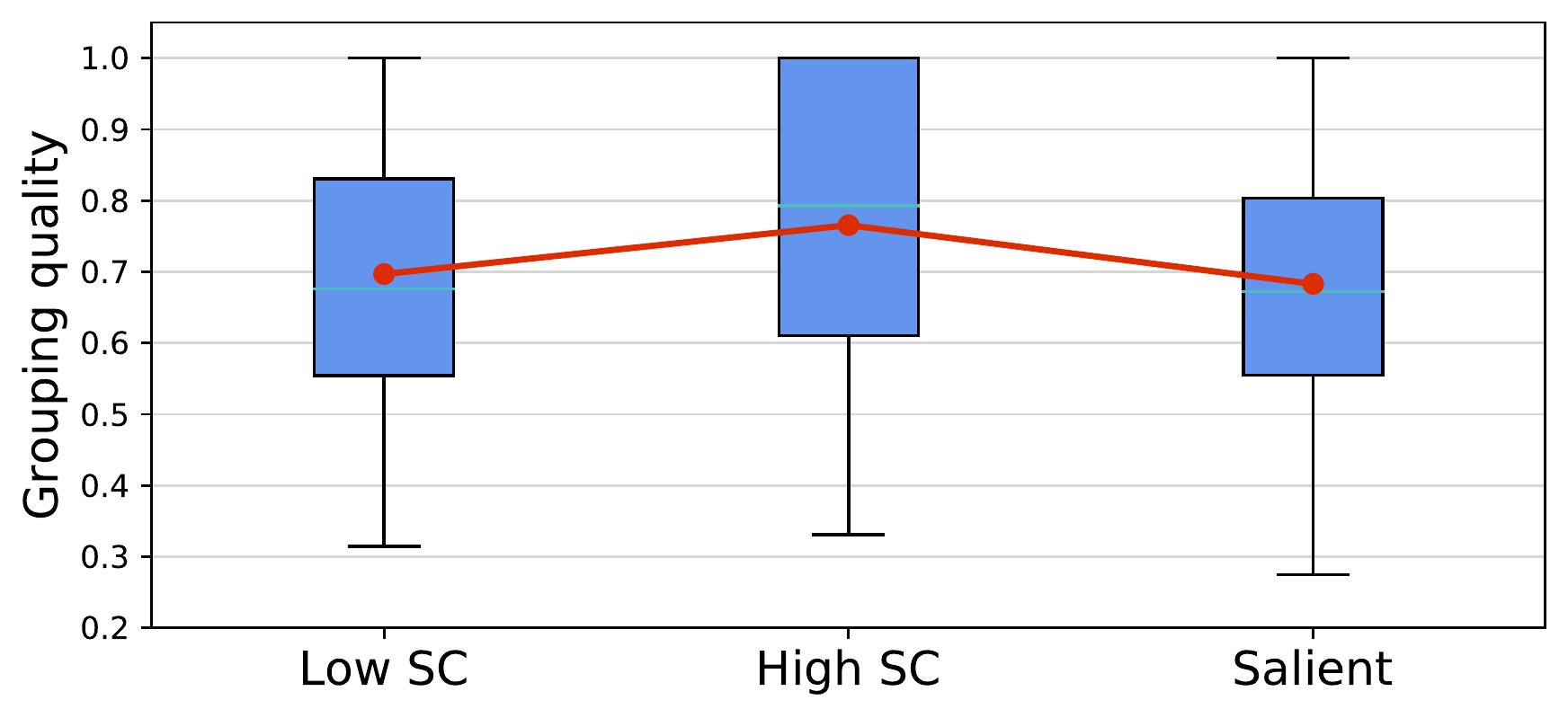}
    \caption{\change{Statistical results of grouping quality in the user study on the class separation measure, where we compare the user performance on star glyph grouping when the glyphs are given in different coordinate orders based on low and high silhouette coefficients and salient features. We see that users can tell the perceptual difference between different glyphs more clearly (higher grouping quality) for sets ordered with a high $SC$. }
     }
	\label{fig:user_study}
\end{figure}

%% file: figures/failure_case.tex
\begin{figure}[tb]
    \centering
    \includegraphics[width=0.92\linewidth]{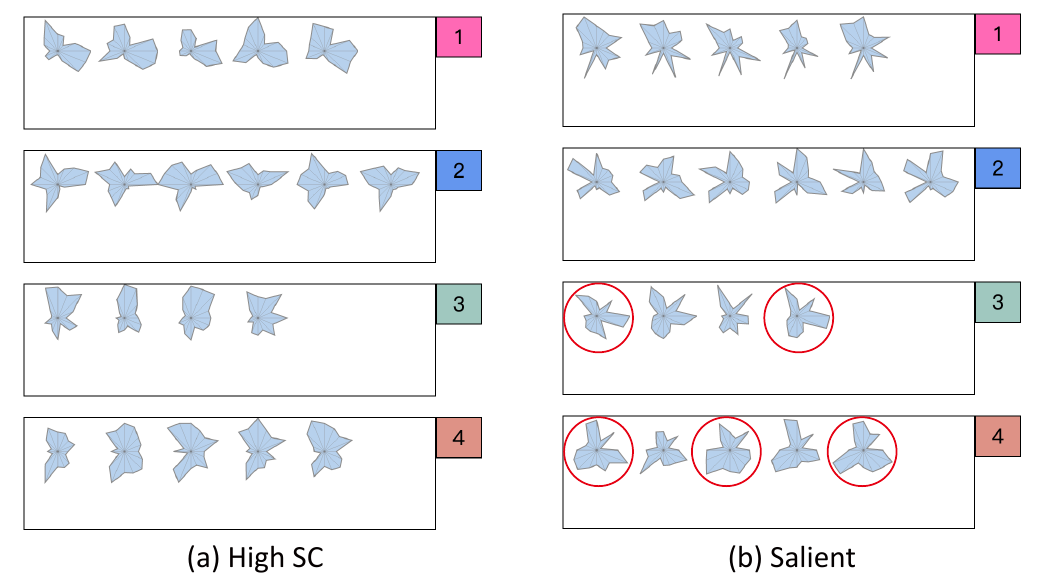}
	\caption{\change{The dataset where the method \emph{Salient} obtains a  higher grouping quality than \emph{High SC}. With the order provided by \emph{High SC}, most of the users are unable to tell Class 3 and Class 4 apart and tend to combine them into one set, which results in 3 groups and a lower grouping quality in the end. The results for \emph{Salient} display a mix between these two classes, but mostly result in a grouping into 4 classes, which leads to a higher grouping quality.
	The star glyphs that are frequently put by users into one group are highlighted using the red circles.}} 
	\label{fig:failure}
\end{figure}

%% file: figures/tab_comp.tex
\begin{table}[tb]%
	\caption{Comparison of our method to a baseline method using synthetic and real data with set size $m=8$, data dimension $n=16$, and class number $K=2$. The numbers denote the silhouette coefficient ($SC$), computed on the entire dataset, where higher values are better as they indicate a better class separation.}
	\label{tab:comp}
	\begin{minipage}{\columnwidth}
		\begin{center}
		\begin{tabular}{c||c|c|c}
			\hline
		Dataset	&  Input & Baseline~\cite{peng2004clutter} &   Ours \\ \hline
		$Synthetic$ & 0.564 & 0.641 & 0.672 \\ \hline
		$Real$ & 0.386 & 0.461 & 0.478 \\ \hline
		\end{tabular}
		\end{center}
	\end{minipage}
\end{table}%

%% file: figures/comp_baseline.tex

\begin{figure}[tb]
    \centering
    \includegraphics[width=0.98\linewidth]{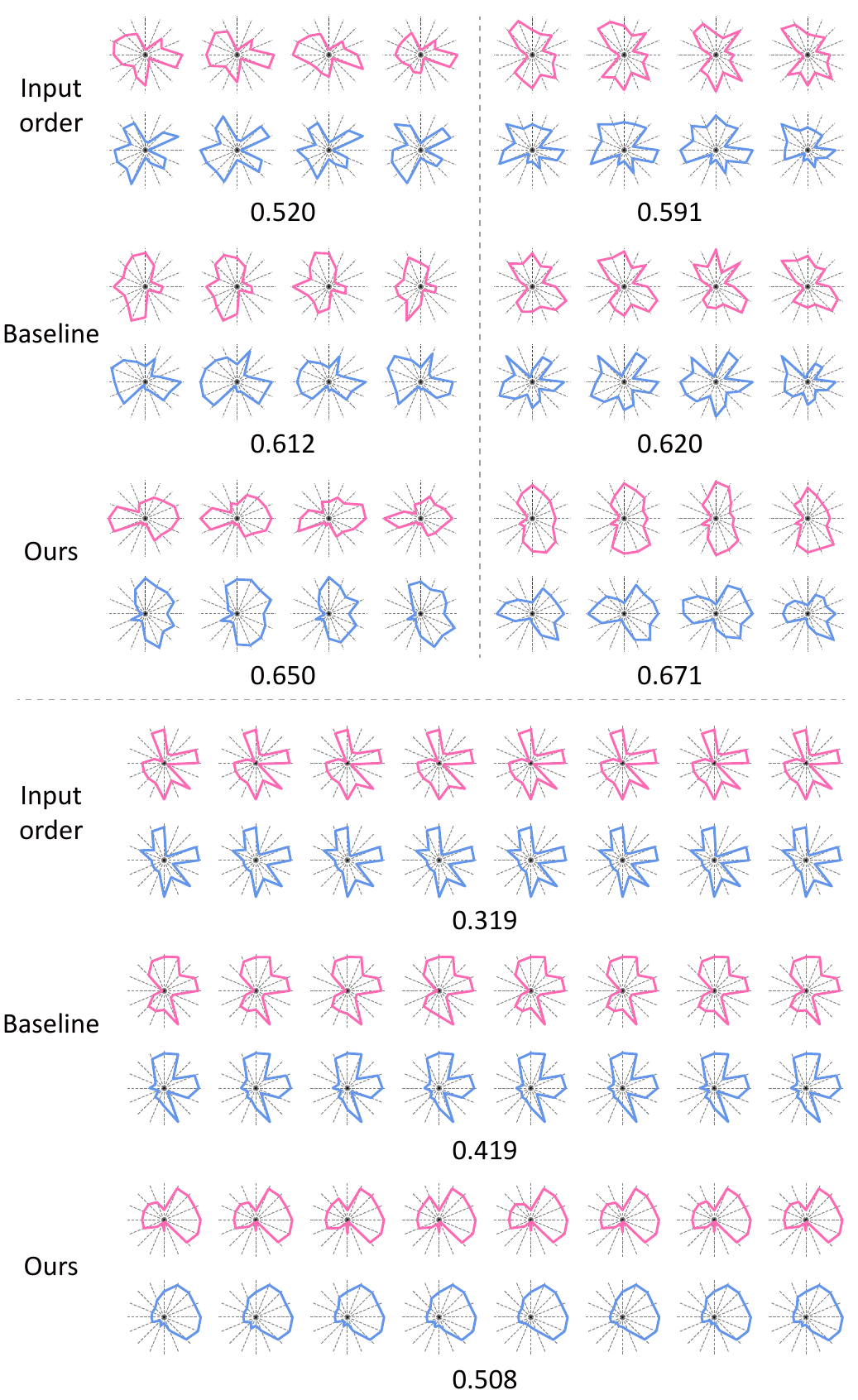}
	\caption{\rh{Comparison to baseline method~\cite{peng2004clutter}: on the top, two examples of synthetic data, and on the bottom, one example of real data. In all examples, glyphs at the same position  represent the same data point but are drawn with different coordinate orders.  Glyph labels are shown by two different colors, and the numbers under each set are the silhouette coefficients ($SC$s) of the orderings. Glyphs from different classes look quite similar when using the input random coordinate order. While the baseline method improves this difference, our ordering provides the best contrast between classes.}}
	\label{fig:comp_baseline}
\end{figure}

%% file: figures/baseline.tex
\begin{figure}[tb]
    \centering
    \includegraphics[width=0.92\linewidth]{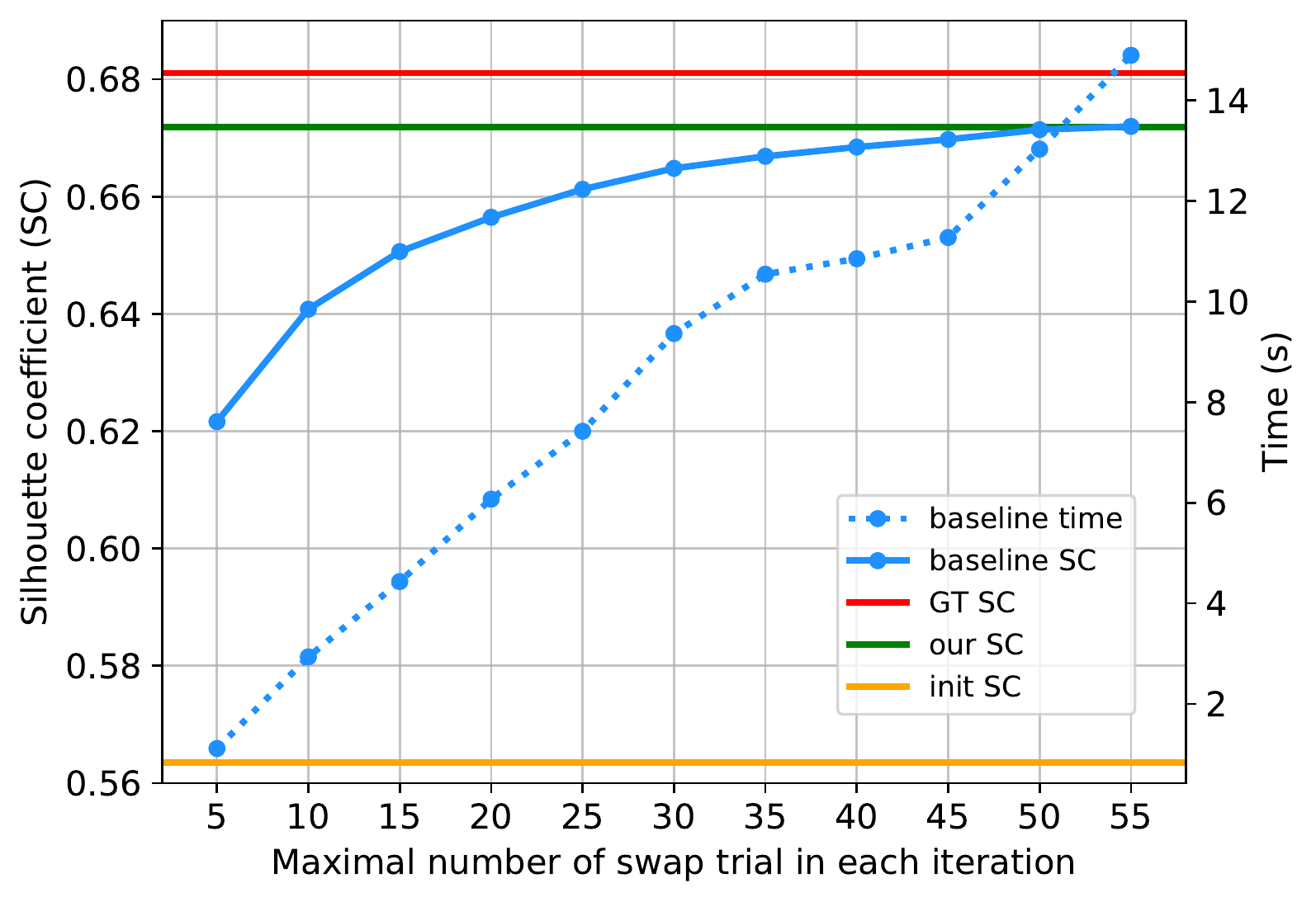}
	\caption{\change{Evaluation of how the performance of the baseline method~\cite{peng2004clutter} changes with the increase of the maximal number of swap trials in each iteration. The baseline (solid blue line) with 55 swaps achieves the same $SC$ as our method, but takes more than 14 seconds (dashed blue line) to perform this computation. In contrast, our method provides the result close to the ground-truth in 0.02s.} 
}         
	\label{fig:baseline}
\end{figure}

%% file: figures/diff_n.tex
\begin{figure}[t]
    \centering
    \includegraphics[width=0.96\linewidth]{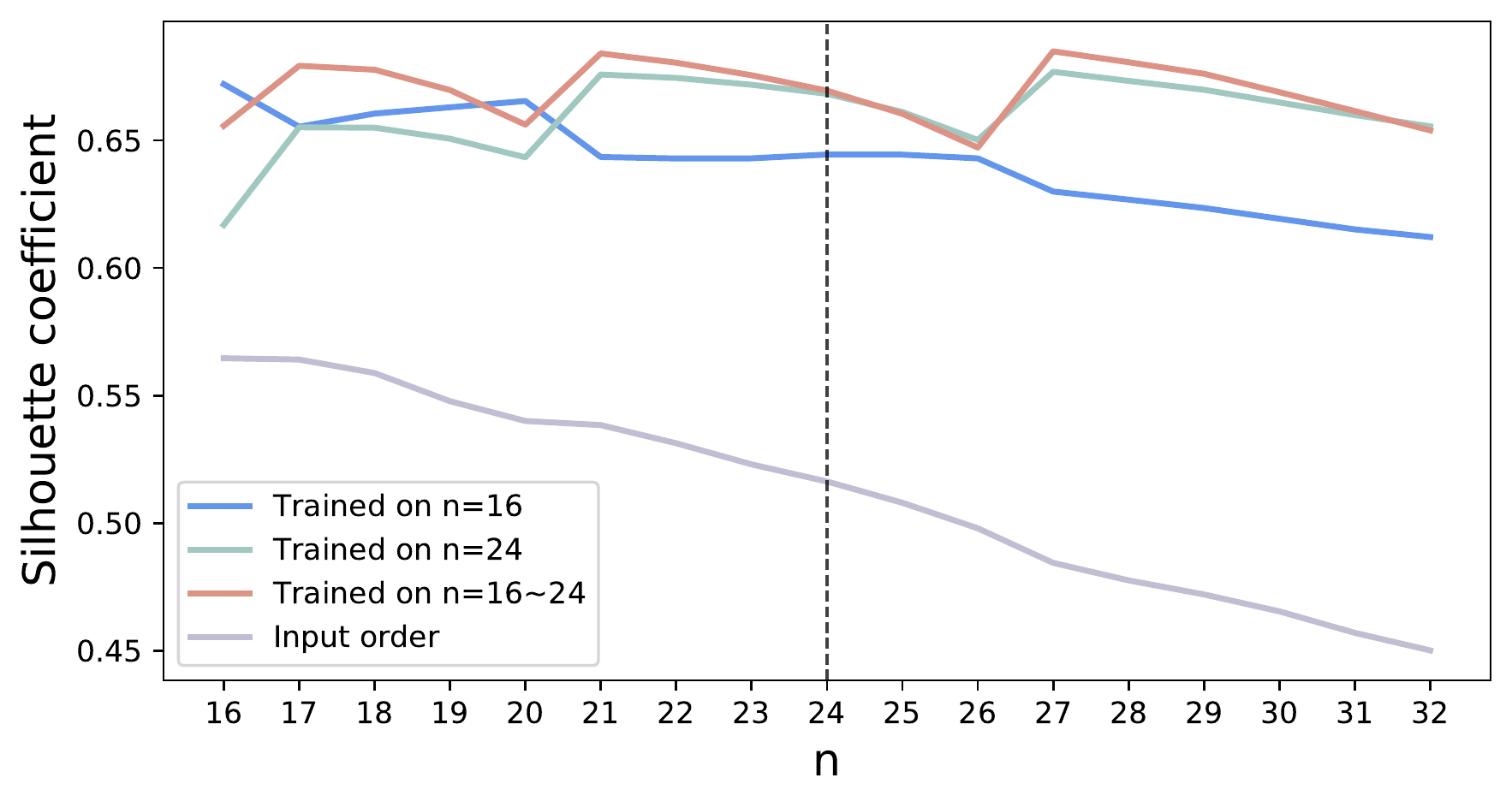}
	\caption{Silhouette coefficients computed on test sets optimized by networks trained with different number of dimensions $n$ (listed in the legend). The $x$-axis denotes the dimensionality of the test set.}
	\label{fig:diff_n}
\end{figure}

%% file: figures/star_result_n.tex
\begin{figure*}[tb]
    \centering
    \includegraphics[width=0.95\linewidth]{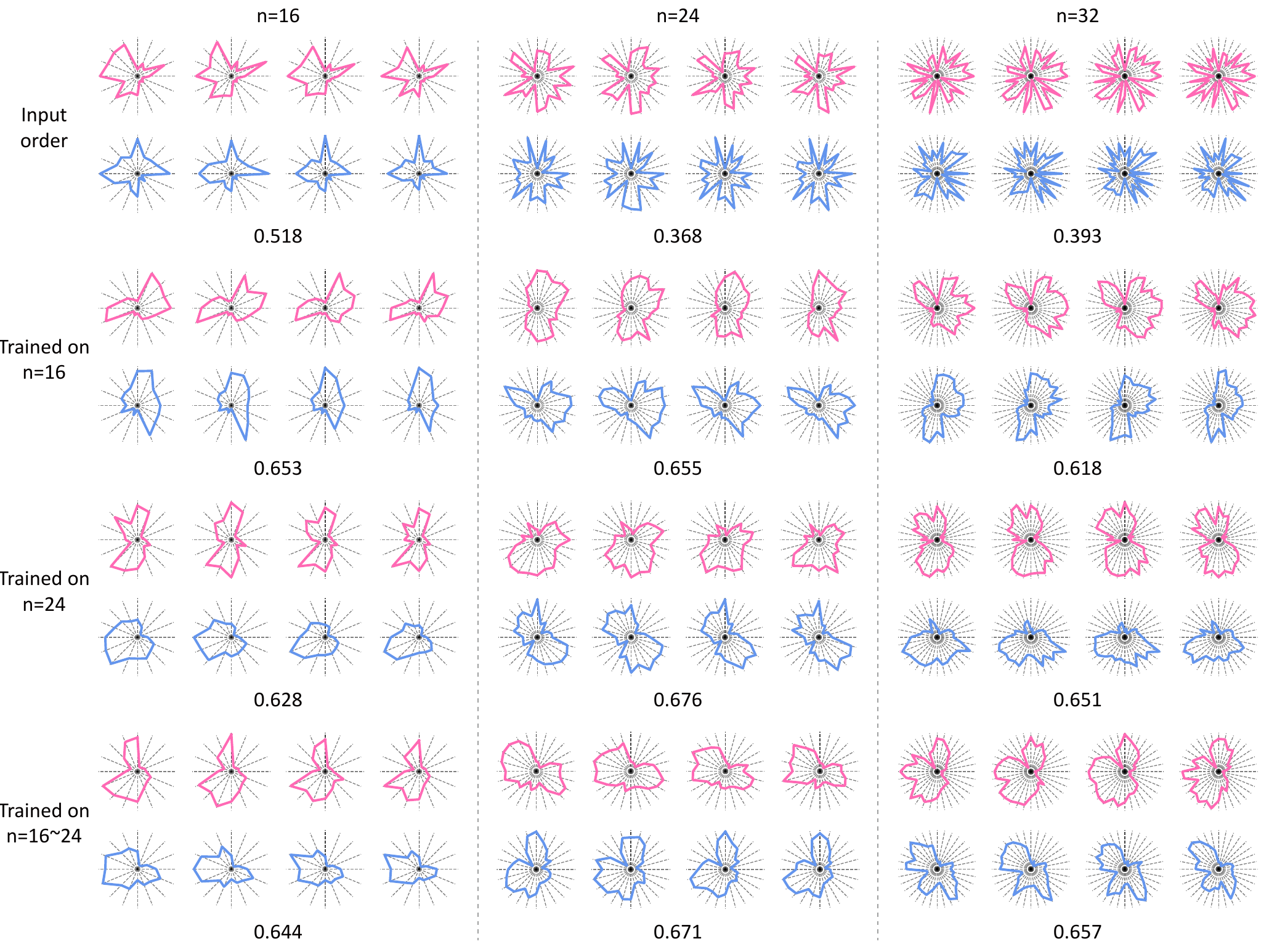}
	\caption{Comparison of coordinate ordering results obtained on test data with networks trained with different number of dimensions $n$. \change{With the random input order, the difference between shapes from different classes is subtle, especially when the data dimension $n$ increases. However, after optimizing the coordinate order of the glyphs with our method, the class difference becomes much more prominent. When comparing results obtained using networks trained with different data, we see that the network trained on $n=16\sim24$ consistently provides good results for data sets with different $n$.} \change{In the example with $n=32$, we can see that with the random input order, the star glyphs have too many spikes which make it hard to tell the difference between shapes from different classes, although the location of the spikes differs. With the order optimized by our method, we see that the prominence of the spikes is reduced and we can focus more on the global shapes of the star glyphs.}}
	\label{fig:star_result_n}
\end{figure*}

%% file: figures/star_result_m.tex
\begin{figure*}[tb]
    \centering
    \includegraphics[width=0.95\linewidth]{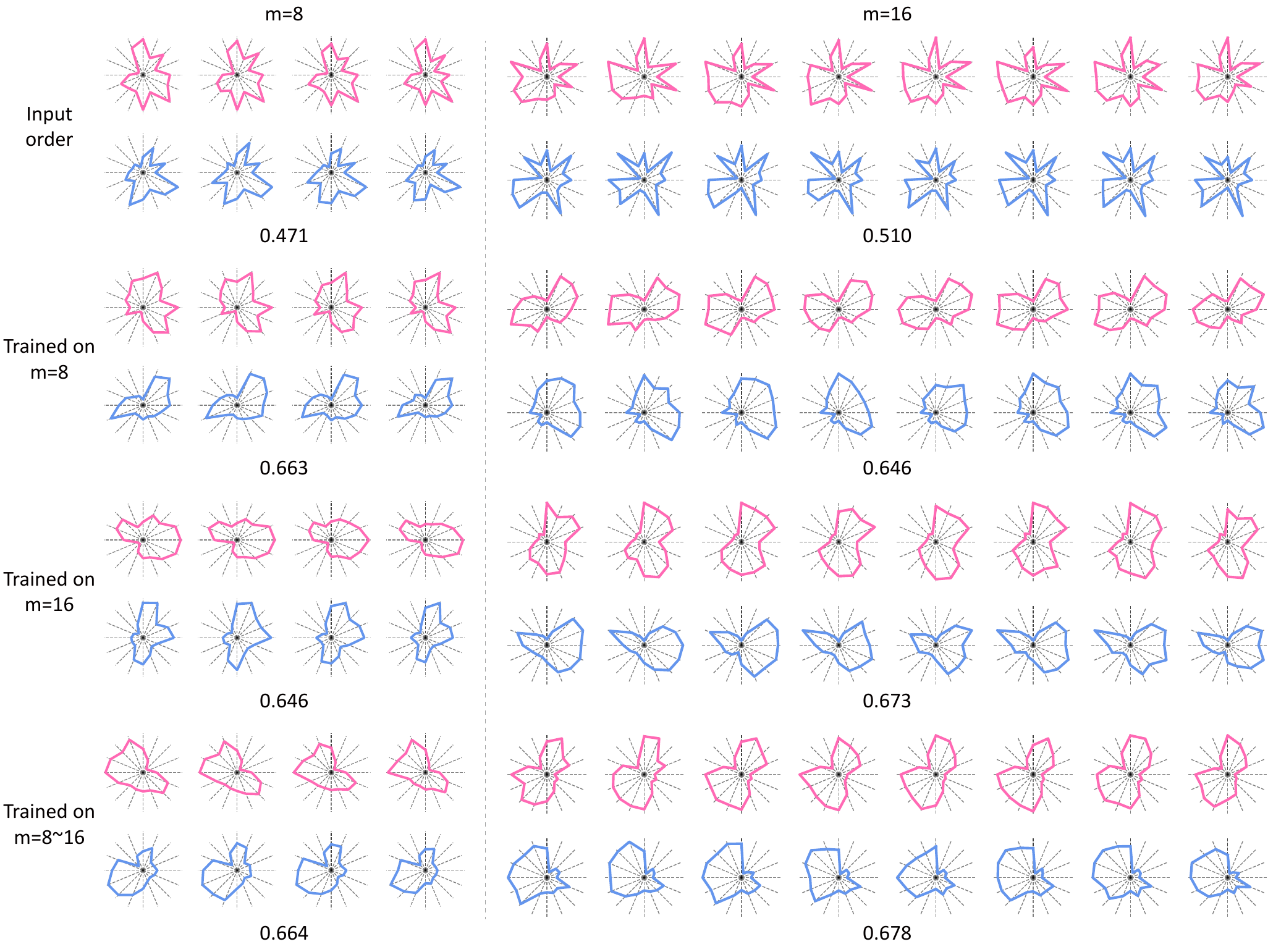}
	\caption{Comparison of coordinate ordering results obtained on test data with networks trained with different number of glyphs $m$. \change{Similar to the results shown in Fig.~\ref{fig:star_result_n}, with random order, it is difficult to tell apart the shapes from different classes. However, with an optimized axis order, the difference between different classes and the similarity within the classes are much clearer.} \change{Moreover, we can see that the networks trained on $m=8$ or $16$ can get close to the best results on data with the same number of glyphs, while the network trained on $m=8\sim16$ provides consistently good results for different $m$.}}
	\label{fig:star_result_m}
\end{figure*}

%% file: figures/diff_m.tex
\begin{figure}[tb]
    \centering
    \includegraphics[width=0.96\linewidth]{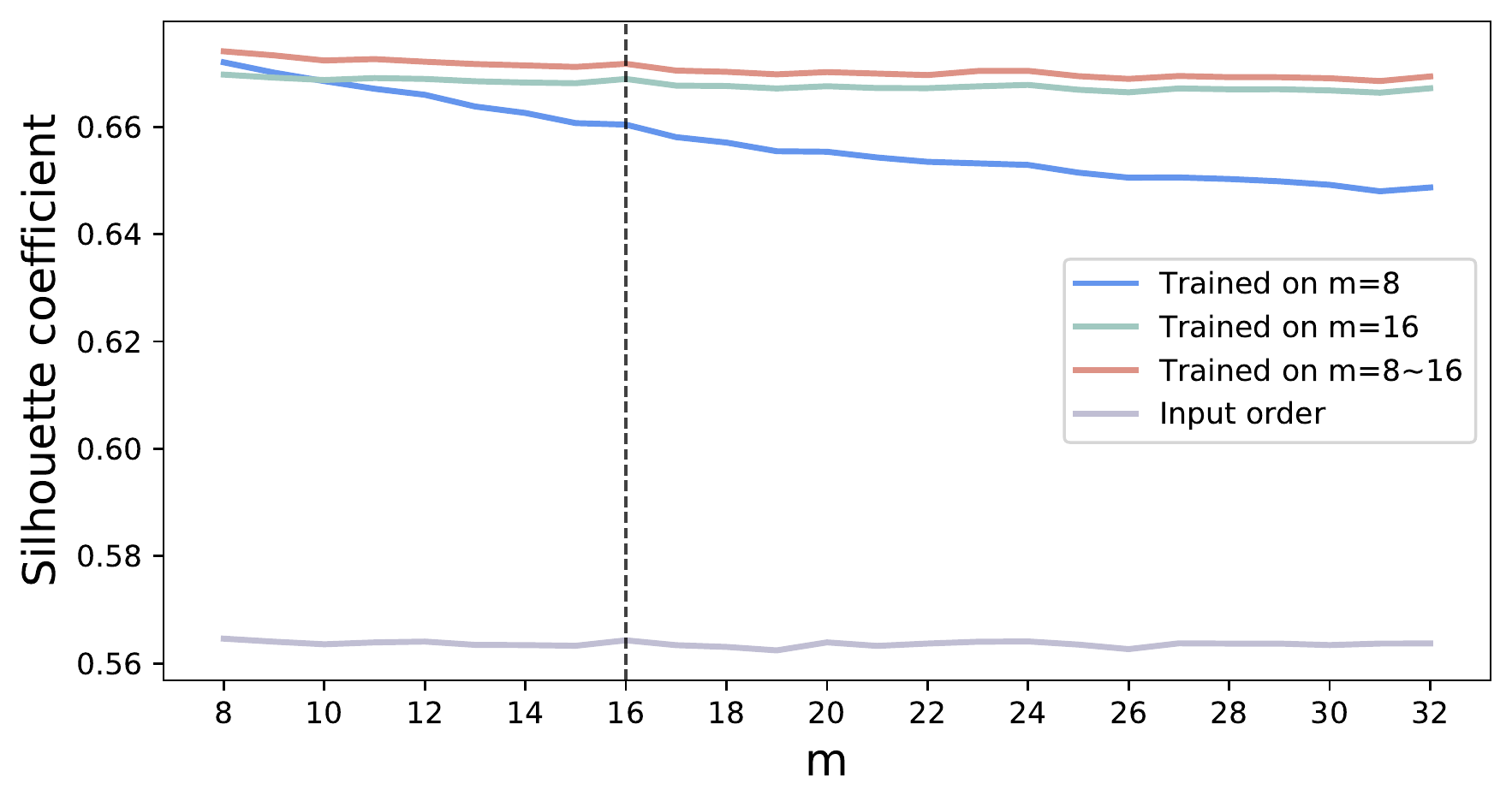}
    \caption{Silhouette coefficients computed on test sets optimized by networks trained with sets of different sizes $m$.}
    \label{fig:diff_m}
\end{figure}

%% file: figures/diff_k.tex
\begin{figure}[tb]
    \centering
    \includegraphics[width=0.96\linewidth]{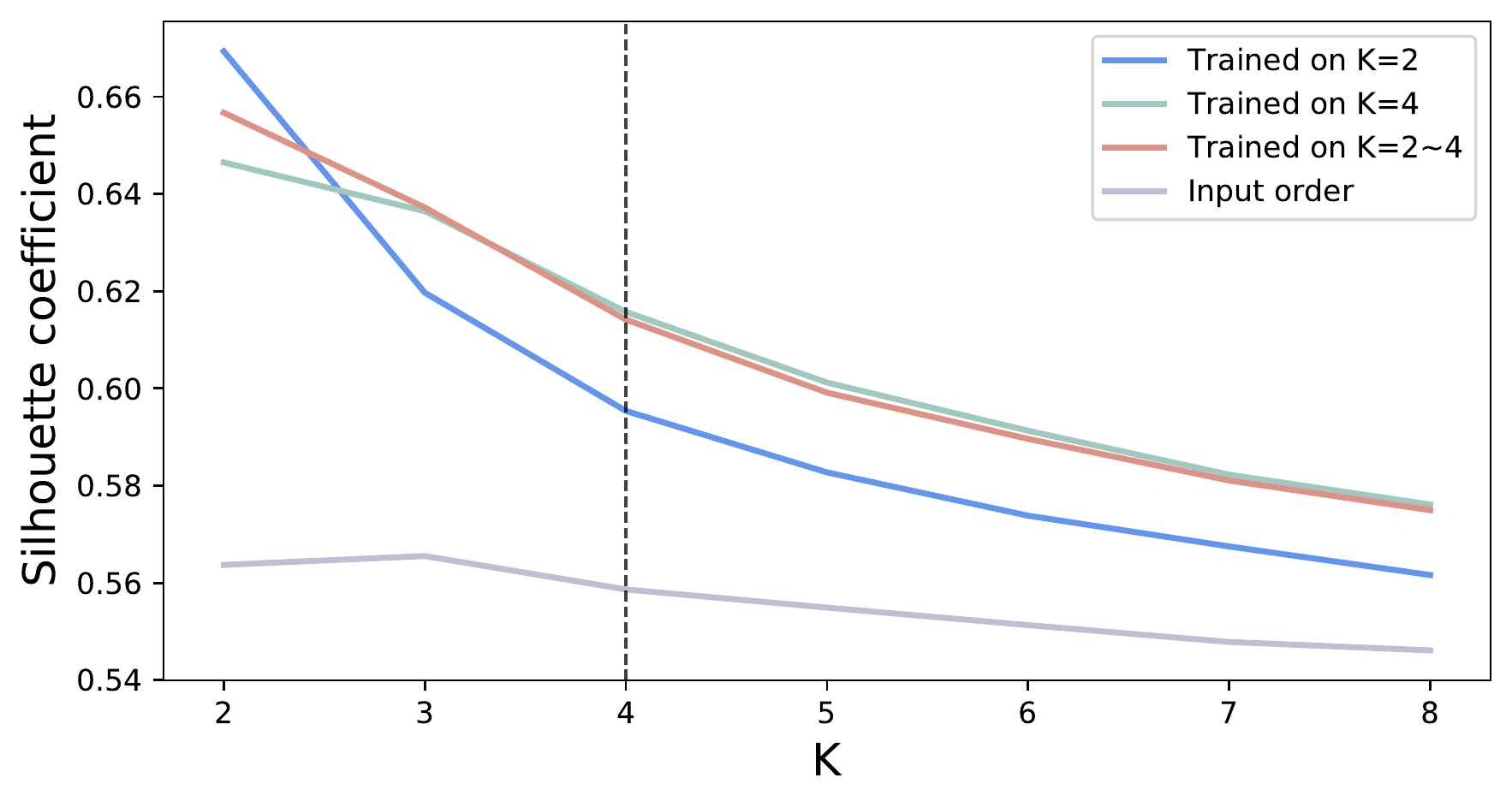}
	\caption{Silhouette coefficients computed on test sets optimized by networks trained with different number of classes $K$. }
	\label{fig:diff_k}
\end{figure}

%% file: figures/star_result_k.tex
\begin{figure*}[tb]
    \centering
    \includegraphics[width=0.98\linewidth]{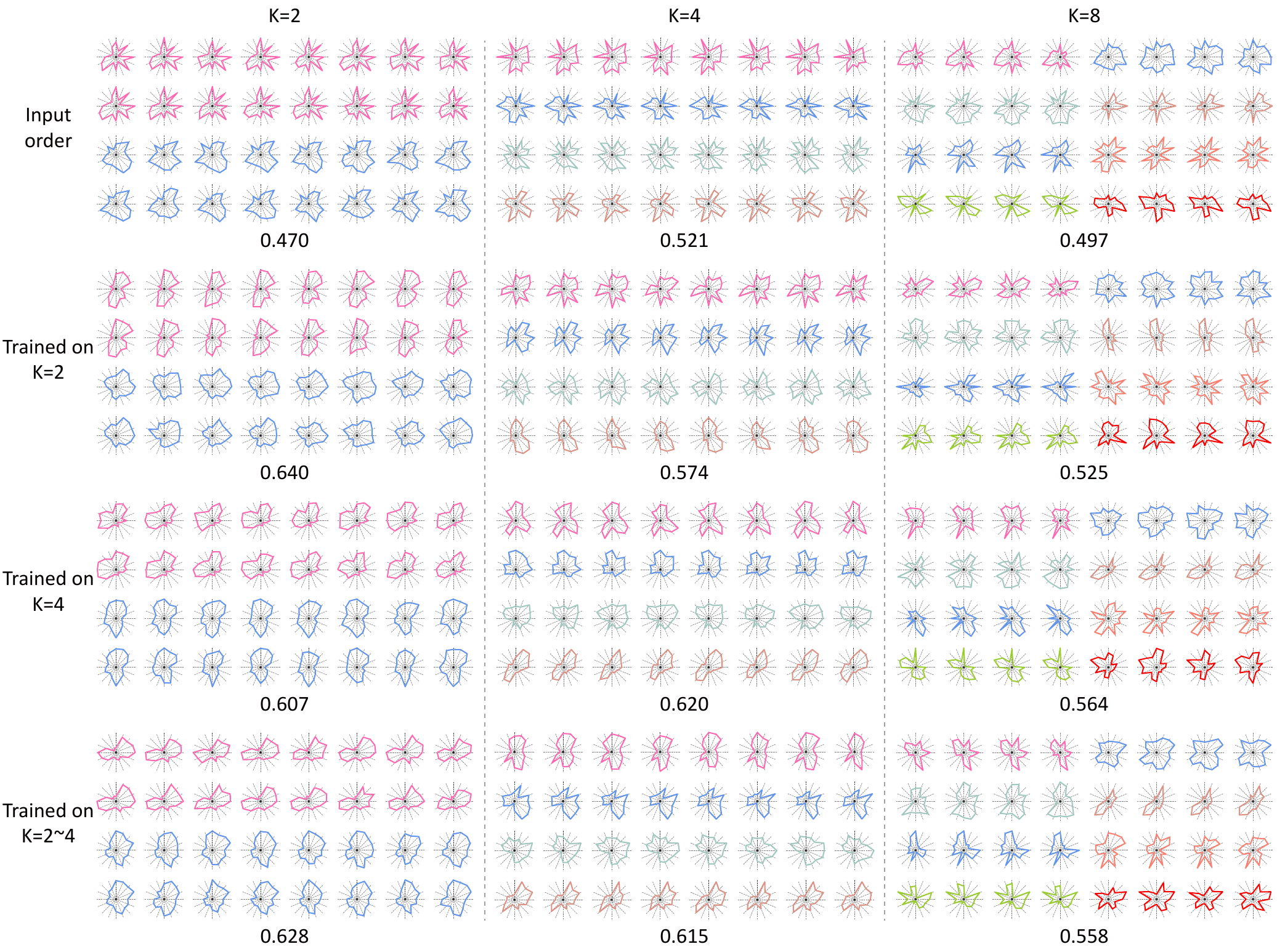}
	\caption{Comparison of coordinate ordering results obtained on test data with networks trained with different number of classes $K$. \change{When the number of classes is small, although there are some differences in the $SC$s, it is still relatively easy to see the difference between shapes of different classes for results obtained using different networks. However, when the number of classes increases, the results obtained with the network trained with either $K=2$ or $K=2\sim4$ show clearer differences, especially for the shapes of the first two classes.}}
	\label{fig:star_result_k}
\end{figure*}

%% file: figures/radviz.tex
\begin{figure*}[tb]
    \centering
    \includegraphics[width=0.98\linewidth]{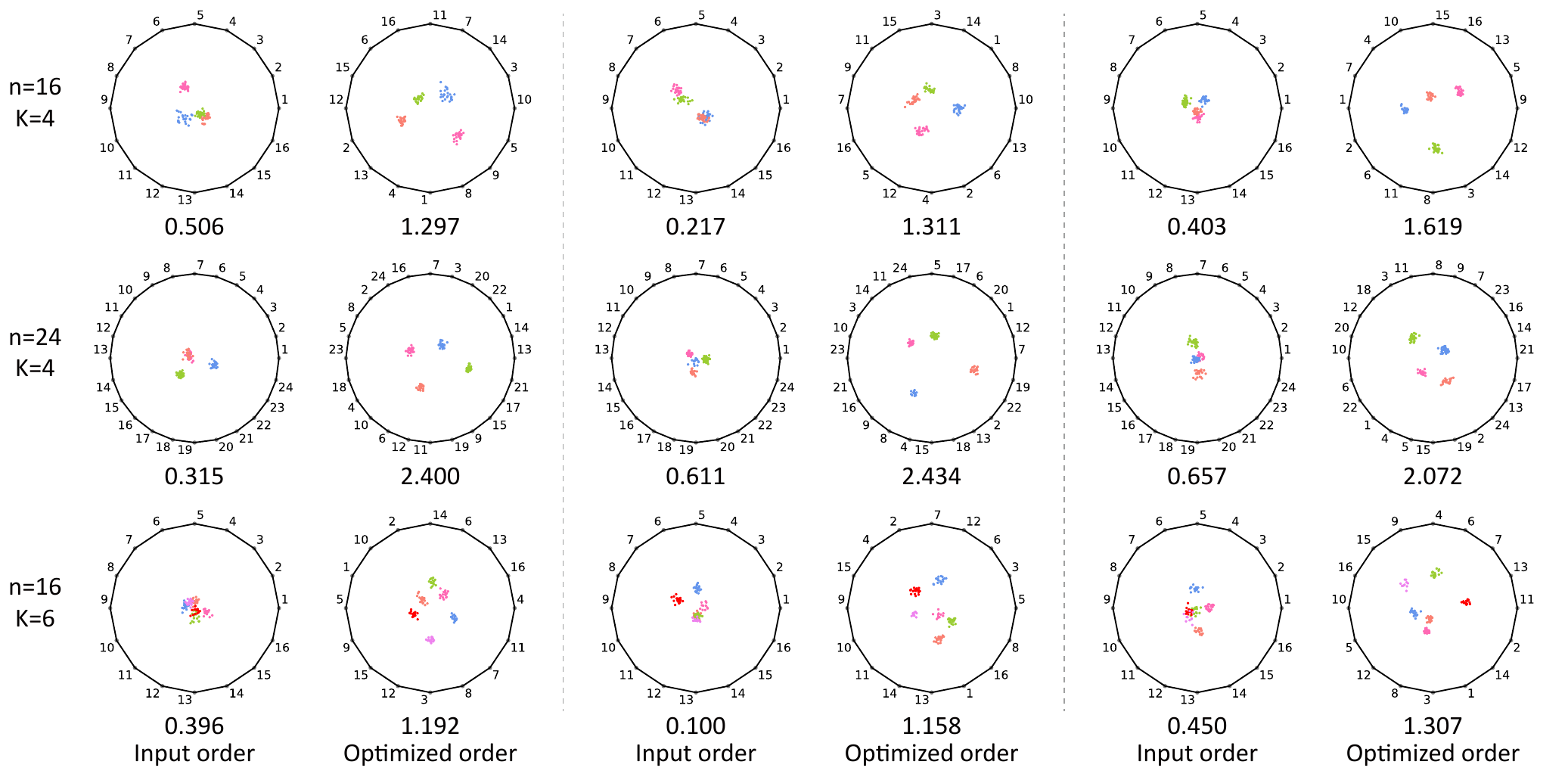}
	\caption{Coordinate optimization results on RadViz, where the
        test sets have different number of dimensions $n$ and different
        number of classes $K$. The number below each plot indicates the 
        Davies-Bouldin index ratio of the plot. \change{We see that the separation between classes is much clearer with the optimized coordinate order.
}}
	\label{fig:radviz}
\end{figure*}

%% file: figures/comp_radviz.tex
\begin{figure}[tb]
    \centering
    \includegraphics[width=0.98\linewidth]{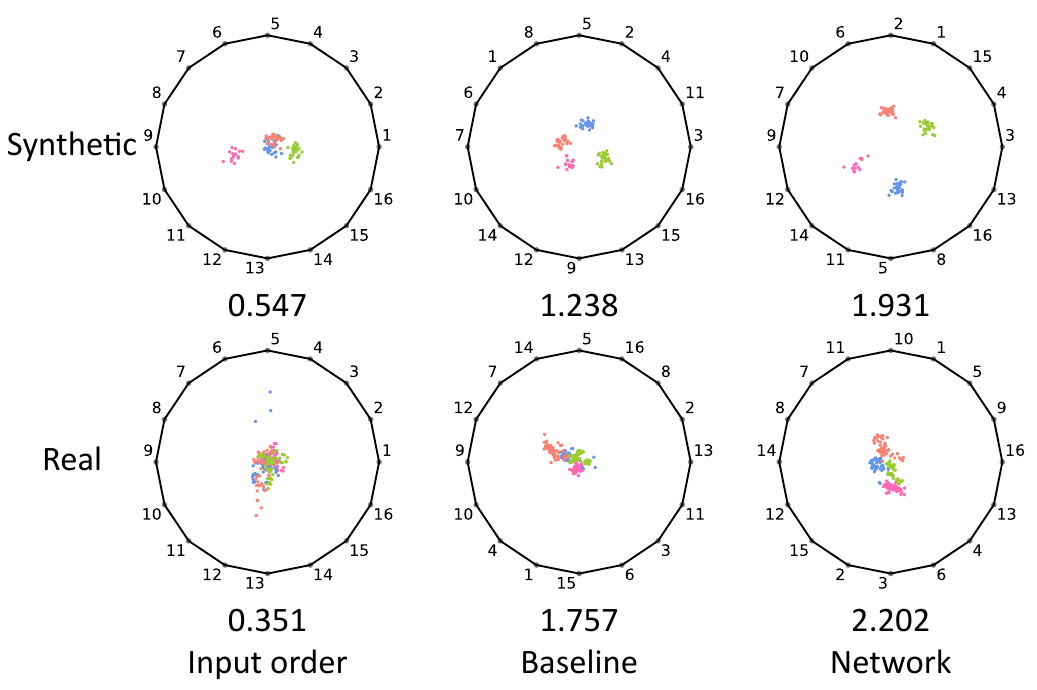}
	\caption{Comparison between RadViz results obtained using our network and a baseline method~\cite{peng2004clutter} on both synthetic and real data. \change{We see that the clusters are distributed further away in the optimized coordinate order than in the order obtained with the baseline method.}}
	\label{fig:comp_radviz}
\end{figure}

%% file: figures/tab_comp_radviz.tex
\begin{table}[!t]%
	\caption{Comparison of our method to a baseline on RadViz using both synthetic and real data in the setting where $m=100$, $n=16$, and $K=4$. The numbers denote the Davies-Bouldin index ratio computed on the entire dataset, where higher values are better.}
	\label{tab:comp_radviz}
	\begin{minipage}{\columnwidth}
		\begin{center}
		\begin{tabular}{c||c|c|c}
			\hline
		Dataset	&  Input & Baseline~\cite{peng2004clutter} &   Ours \\ \hline
		$Synthetic$ & 0.547 & 1.394 & 1.701 \\ \hline
		$Real$ & 0.555 & 1.375 & 1.723 \\ \hline
		\end{tabular}
		\end{center}
	\end{minipage}
\end{table}%

%% file: conclusion.tex
\section{Conclusion}
\label{sec:conclusion}

We introduced a method to optimize the coordinate ordering of sets of star glyphs associated with multiple class labels, in order to maximize perceptual class separation. Our method measures the class separability using a silhouette coefficient based on visual differences between glyphs derived from a shape context descriptor. The coefficient is then used to optimize ordering axis of the glyphs with a set-to-sequence neural network. We showed that training the network with a variety of data allows our method to optimize the coordinate ordering for sets of glyphs with varying size, number of dimensions and classes. In addition, we demonstrated with two user studies that the orders provided by our method are preferred by users for revealing class separation. Finally, we  showed that our method is general enough to be applied to other types of plots such as RadViz.

\paragraph{Limitations and future work.}
In our current setting, we use the silhouette coefficient on the basis of shape context distances as the objective function to guide coordinate ordering. However, other metrics 
can be used. It would be interesting to investigate other reward functions for tasks beyond class separation.

\od{
Moreover, our method can be applied to input datasets with different sizes, dimensions, and number of classes. However, if these parameters differ significantly from the training data, the quality of the results may be low and the network may need to be trained with datasets of more similar parameters. An interesting future direction would therefore be to investigate ways of designing  networks so that they are able to deal with datasets with a variety of parameters. 

Another point is that the datasets that we used to train the network were generated using Gaussian distributions, while real data may incorporate different distributions. Thus, the policy we use to train from synthetic datasets may not fit well to real data. It would be interesting to explore ways to simulate real data distributions more faithfully. 
}
As mentioned in Section~\ref{sec:architecture}, we use an RNN to encode the input data in order to handle datasets with different sizes. So, in theory, different data orders may lead to different results. Although we only found stable results in our experiments, it would be worthy to explore input encodings that are invariant w.r.t. to data order. 

Currently, our method works well on star glyphs and RadViz, since the quality measures that guide coordinate ordering capture the global shape (features) of these two charts. For other types of plots such as parallel coordinates, local features such as line crossings between two adjacent axes are visually more important. Our method may not work well in these cases since it is focused on global features. Thus, it would be interesting to design more suitable metrics and network architectures to optimize local features for axis ordering in, e.g., parallel coordinates.
